\documentclass[letterpaper]{article} % DO NOT CHANGE THIS
\usepackage{aaai2026}  % DO NOT CHANGE THIS
\usepackage{times}  % DO NOT CHANGE THIS
\usepackage{helvet}  % DO NOT CHANGE THIS
\usepackage{courier}  % DO NOT CHANGE THIS
\usepackage[hyphens]{url}  % DO NOT CHANGE THIS
\usepackage{amsfonts}
\usepackage{graphicx} % DO NOT CHANGE THIS
\usepackage{natbib}  % DO NOT CHANGE THIS AND DO NOT ADD ANY OPTIONS TO IT
\usepackage{caption} % DO NOT CHANGE THIS AND DO NOT ADD ANY OPTIONS TO IT
\newcommand{\ourwork}{\textit{NutriScreener}}
\usepackage{booktabs}
\usepackage{multirow}
\usepackage{subcaption}
\usepackage{amsmath}
\usepackage{algorithm}
\usepackage{algorithmic}
\usepackage{xurl}
\usepackage{newfloat}
\usepackage{listings}
\usepackage[table]{xcolor}   % <-- the “table” option is crucial
\DeclareCaptionStyle{ruled}{labelfont=normalfont,labelsep=colon,strut=off} % DO NOT CHANGE THIS
\floatstyle{ruled}
\newfloat{listing}{tb}{lst}{}
\floatname{listing}{Listing}
\nocopyright
\title{NutriScreener: Retrieval-Augmented Multi-Pose Graph Attention Network for Malnourishment Screening}

\author {
    Misaal Khan\textsuperscript{\rm 1,\rm 2},
    Mayank Vatsa\textsuperscript{\rm 1},
    Kuldeep Singh\textsuperscript{\rm 2},
    Richa Singh\textsuperscript{\rm 1}
}

\affiliations {
    \textsuperscript{\rm 1}Indian Institute of Technology Jodhpur, Rajasthan, India\\
    \textsuperscript{\rm 2}All India Institute of Medical Sciences Jodhpur, Rajasthan, India\\
    \{khan.9, mvatsa, richa\}@iitj.ac.in, singhk@aiimsjodhpur.edu.in\\
}

\usepackage{bibentry}
\begin{document}

\maketitle

\begin{abstract}
Child malnutrition remains a global crisis, yet existing screening methods are laborious and poorly scalable, hindering early intervention. In this work, we present \ourwork{}, a retrieval-augmented, multi-pose graph attention network that combines CLIP-based visual embeddings, class-boosted knowledge retrieval, and context awareness to enable robust malnutrition detection and anthropometric prediction from children's images, simultaneously addressing generalizability and class-imbalance. In a clinical study, doctors rated it 4.3/5 for accuracy and 4.6/5 for efficiency, confirming its deployment readiness in low-resource settings. Trained and tested on 2,141 children from AnthroVision and additionally evaluated on diverse cross-continent populations, including ARAN and an in-house collected CampusPose dataset. It achieves 0.79 recall, 0.82 AUC, and significantly lower anthropometric RMSEs, demonstrating reliable measurement in unconstrained, pediatric settings. Cross-dataset results show up to 25\% recall gain and up to 3.5 cm RMSE reduction using demographically matched knowledge bases. \ourwork{} offers a scalable and accurate solution for early malnutrition detection in low-resource environments.
\end{abstract}

% Uncomment the following to link to your code, datasets, an extended version or similar.
% You must keep this block between (not within) the abstract and the main body of the paper.
\textbf{Toolkit: } \url{https://www.iab-rubric.org/resources/healthcare-datasets/nutriscreener}

% \texttt{https://www.iab-rubric.org/\discretionary{}{}{}resources/\discretionary{}{}{}healthcare-datasets/\discretionary{}{}{}nutriscreener}

% \begin{links}

% \textbf{Toolkit: } \url{https://www.iab-rubric.org/resources/healthcare-datasets/nutriscreener}    

% % \link{Datasets}{https://aaai.org/example/datasets}
%     % \link{Extended version}{to be updated}
% \end{links}

\section{Introduction}
% Malnutrition remains a pervasive and often underdiagnosed global health crisis, affecting all age groups but especially impacting children in resource-constrained regions, leading to severe long-term health complications, increased morbidity, and mortality~\cite{khan2022ai}.

As of 2024, approximately 150 million children under five years of age globally suffer from stunting, and over 42 million from wasting, both direct outcomes of chronic and acute malnutrition. These conditions are recognised by the World Health Organization as leading causes of irreversible developmental impairment and mortality in early childhood\footnote{ \url{https://www.who.int/data/gho/data/themes/topics/joint-child-malnutrition-estimates-unicef-who-wb}, last viewed: Aug. 1, 2025}~\cite{who_malnutrition_2024}. 
% worldhunger_child_2022

Malnutrition remains a pervasive and underdiagnosed global health crisis, especially in children, leading to long-term developmental issues and increased morbidity. These consequences are further magnified in low-resource settings where timely screening is often inaccessible~\cite{khan2022ai}. Conventional assessments rely on manual anthropometric measurements, such as mid-upper arm circumference (MUAC) tapes, weight-for-height charts, and questionnaires, which are labor-intensive, error-prone, and often delayed, making them unsuitable for rapid or scalable deployment~\cite{lee_impact_2025, janssen2024artificial, khan2023nutriai}. There is a pressing need for automated, reliable malnutrition screening that minimizes reliance on manual measurements.

% Malnutrition remains a pervasive and underdiagnosed global health crisis, particularly affecting children and contributing to long-term developmental impairments, increased morbidity, and mortality worldwide. These consequences are further magnified in resource-constrained regions, where limited access to timely screening and intervention further hinders early detection and treatment~\cite{khan2022ai}. Additionally, malnutrition significantly increases healthcare costs and disease burden in adult populations globally~\cite{lee_impact_2025}. Traditional assessment methods rely on predefined questionnaires and manual anthropometric measurements, such as mid-upper arm circumference (MUAC) tapes and weight-for-height charts, which rely on manual measurements and direct patient interaction, which are labor-intensive, prone to error, and subject to subjective interpretation, and often delay timely interventions making them impractical in low-resource or emergency settings~\cite{janssen2024artificial, khan2023nutriai}. Therefore, efficient and reliable screening tools are urgently needed to enable early, automated malnutrition detection without relying on manual assessments.

\begin{figure}[!t]
    \centering
    \includegraphics[width=1\linewidth]{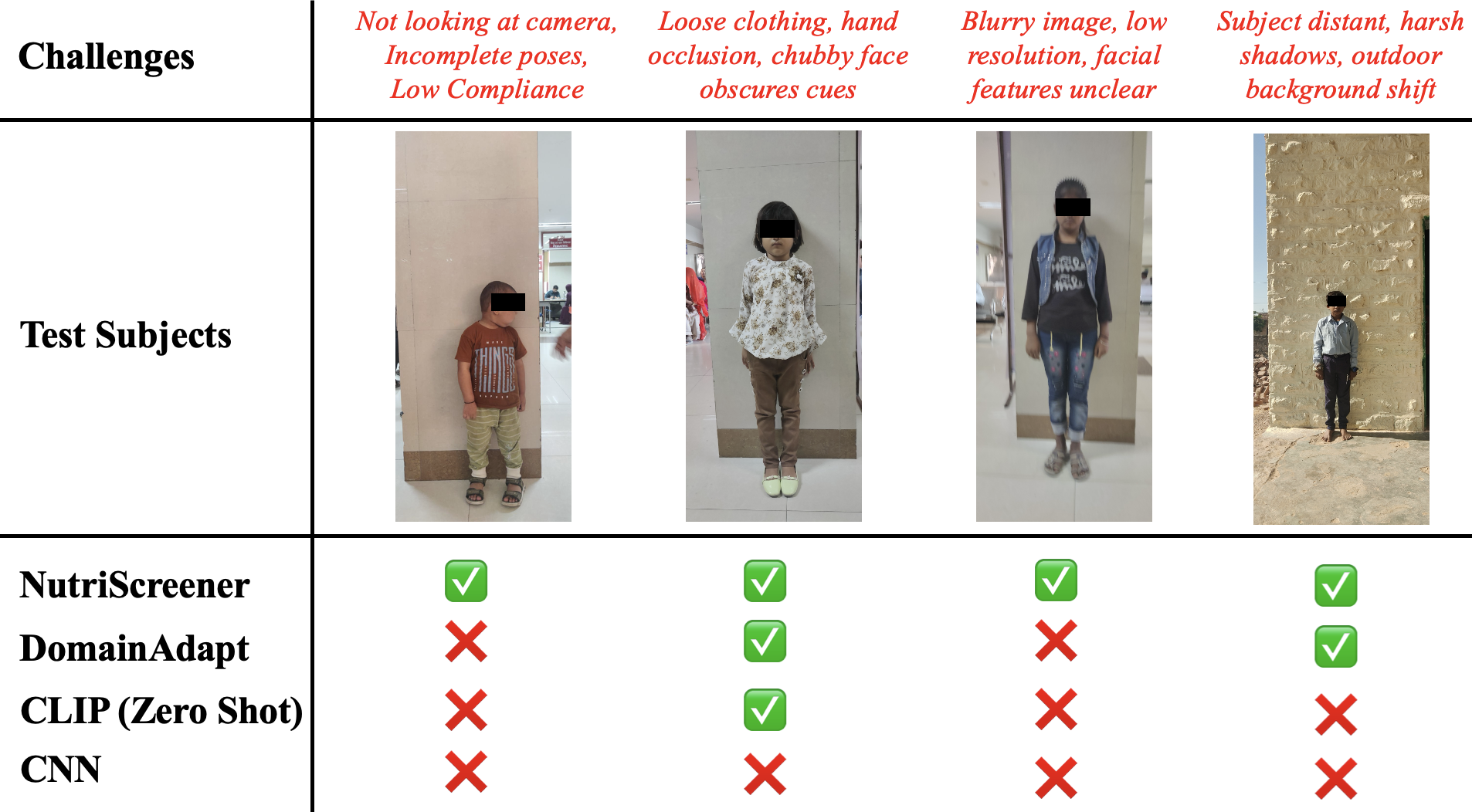}
    \caption{Robustness of \ourwork{} to detect \textit{malnourishment} across challenging real-world malnutrition scenarios.}
    \label{fig:placeholder}
\end{figure}

Prior studies have explored malnourishment detection using facial and full-body imagery, but most facial-image approaches~\cite{wang2023establishing, tay2022use} target elderly populations and do not generalize to children. Solutions like Microsoft's Child Growth Monitor~\cite{childgrowthmonitor2025} require specialized hardware (infrared depth sensors), limiting accessibility in high-need regions. Most AI models~\cite{janssen2024artificial} remain experimental and lack validation for routine clinical use, leaving traditional anthropometry as the clinical standard. Current models~\cite{aanjankumar2025prediction, khan2024domainadapt}, trained on limited datasets, often exhibit majority-class bias, reducing sensitivity to malnourished cases. As no single viewpoint captures all diagnostic cues, robust screening demands integration across multiple image poses.

Motivated by these insights, we propose \ourwork, a multi-pose malnutrition assessment framework that integrates Graph Attention Networks (GATs)\cite{veličković2018graph} with retrieval-augmented learning for robust classification and anthropometric estimation. Each subject is modeled as a graph whose nodes contain pose-wise embeddings extracted using the Contrastive Language–Image Pre-training (CLIP) image encoder \cite{radford2021learning}, augmented with age metadata. We adopt CLIP due to its strong cross-domain generalization, with prior work showing that its pretrained semantic features implicitly encode fine-grained anatomical and morphological cues relevant to medical imaging~\cite{yu2024cp}, anthropometry~\cite{khandelwal2024nurturenet}, and few-shot learning~\cite{li2025text}. Graph Attention Network (GAT) layers then model inter-pose relationships, promoting cross-view consistency for both classification and regression tasks. To address the significant class imbalance characteristic of malnutrition datasets, we construct a population-level Knowledge Base (KB) and retrieve semantically similar samples using (Facebook AI Similarity Search) FAISS~\cite{Douze2024The}. Retrieved exemplars are used to boost minority-class predictions through a retrieval-informed augmentation mechanism. Finally, the retrieval output is adaptively fused with GAT predictions using a learned context-aware fusion layer that dynamically balances retrieval information and GAT confidence scores to improve both precision and recall. The contributions of this work are summarized as follows:

\begin{itemize}
    \item To the best of our knowledge, this is the first effort to integrate retrieval augmentation with multi-pose Graph Attention Networks for malnutrition screening, jointly performing anthropometric regression and binary classification from 2D image inputs.
    
    \item Our framework supports generalization to new populations by requiring only a few representative samples in the KB, addressing severe class imbalance and enabling practical deployment in low-resource settings.

    \item We propose a context-aware gated fusion mechanism to combine GAT and retrieval outputs based on model confidence and local density, enabling robustness under pose variability, domain shifts, and label imbalance.
    
    \item We validate \ourwork{} through a real-world clinician study, confirming its accuracy, efficiency, and deployment readiness. The toolkit, labeled KB and CampusPose dataset are released to support field use and research. 
    
    % Watch Demo \footnote{\url{https://shorturl.at/WaHdZ} (Link to the anonymized demo video, Last Access: August 1, 2025)}.

\end{itemize}

\section{Related Work}
% \subsection{AI Approaches for Malnutrition}
AI-based malnutrition assessment remains underexplored due to limited public datasets and challenges in adapting vision models to low-resource clinical settings~\cite{khan2023nutriai, khan2022ai}. Early work relied on small-scale data: ARAN~\cite{MohammedKhan2025ARAN} includes only 512 Kurdish children from a single region, while AnthroVision~\cite{khan2024domainadapt} contains 2,142 multi-view samples of Indian children, but is geographically restricted to Jodhpur, Rajasthan. A recent multitask model (DomainAdapt)~\cite{khan2024domainadapt} trained on AnthroVision showed improved classification but suffered from low malnourishment recall ($\sim$67\%). These limitations highlight the need for more diverse datasets and tailored methods for robust screening.

% \subsection{Foundation Models and Visual Generalization}
% Foundation models trained on extensive image-text corpora (e.g., CLIP;~\cite{radford2021learning}) have recently demonstrated powerful generalization capabilities across diverse visual tasks. Such vision-language models enable robust transfer to medical imaging and anthropometry contexts. In particular, adaptations like MedCLIP~\cite{wang2022medclip} (trained on unpaired radiology images and reports) integrate medical knowledge into contrastive learning, outperforming prior state-of-the-art on zero-shot prediction, supervised classification, and image–text retrieval in biomedical applications. These results suggest that foundation models provide rich semantic features and resilience to domain shifts, which can be leveraged to predict human anthropometric measurements across diverse populations and conditions.

Foundation models like CLIP~\cite{radford2021learning}, trained on large image–text data, exhibit strong generalization across visual tasks and transfer well to medical domains. MedCLIP~\cite{wang2022medclip}, adapted using unpaired radiology image–report pairs, outperforms prior methods in zero-shot prediction, classification, and retrieval. Such models offer rich semantic features and robustness to domain shifts, making them well-suited for anthropometric prediction across diverse populations.

\begin{figure*}
    \centering
    \includegraphics[width=1\linewidth]{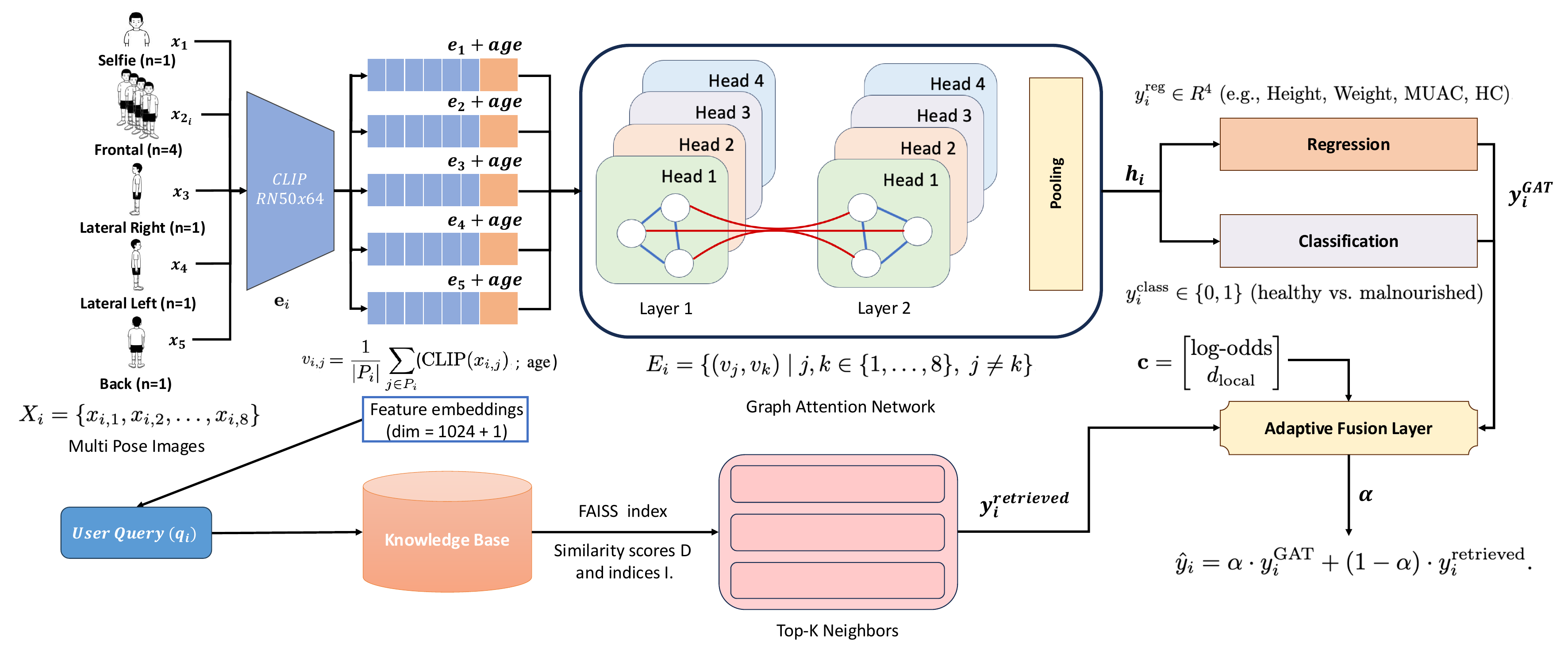}
    \caption{Architectural overview of \ourwork. The system takes a subject’s multi-pose images and age as input. It extracts pose-wise visual embeddings using a CLIP-based encoder and aggregates them, then passes the aggregated embeddings through a graph attention network to produce image-based predictions for both classification and regression tasks. In parallel, the same aggregated embeddings are used to query a FAISS-indexed knowledge base, yielding retrieval-based predictions. An adaptive fusion layer then combines the graph-based and retrieval-based outputs.}
    
    \label{fig:architecture}
\end{figure*}

Graph neural networks (GNNs) and multi-view learning provide structured approaches to model body morphology from images. GNNs represent body parts or landmarks as graph nodes, capturing relational constraints to improve shape inference. For instance, ~\cite{Li2020DMGNN} models multiscale joint relations for realistic pose prediction, while ~\cite{Kolotouros2019GraphCMR} uses graph convolutions on body meshes to regress 3D shape. Multi-view methods further enhance anthropometric estimates by aggregating information across poses. Liu et al.~\cite{Liu2018MultiViewAnthropometry} show that even linear models achieve accurate height and MUAC prediction with multi-angle inputs. Together, GNNs and multi-view consistency improve robustness in image-based anthropometry.

% \subsection{Retrieval-Augmentation for Imbalanced Data}
% To address class imbalance in malnutrition classification, recent methods draw on retrieval-augmented learning. A seminal approach is Retrieval Augmented Classification (RAC), which introduced an external memory index (built with FAISS) to supply nearest-neighbor exemplars during image classification~\cite{Long2022RAC, Liu2025SWAT}. Likewise, COBRA~\cite{Das2025COBRA} selects support examples using a combinatorial mutual information criterion that favors diversity in addition to similarity, yielding significant performance gains for minority classes with negligible extra cost.
% \subsection{Retrieval-Augmentation for Imbalanced Data}
Retrieval-augmented learning addresses class imbalance by incorporating external exemplars during inference. RAC~\cite{Long2022RAC, Liu2025SWAT} introduces a FAISS-based memory index to retrieve nearest neighbors for improved classification. COBRA~\cite{Das2025COBRA} further optimizes retrieval via mutual information to balance similarity and diversity, enhancing recall with minimal overhead.

% Given the scarcity of well-annotated, balanced malnutrition datasets, we hypothesize that CLIP’s pretrained visual representations inherently capture generalizable indicators of malnutrition, such as subtle variations in body proportions, posture, clothing fit, and diverse contextual backgrounds even without task-specific training.
% Our investigation reveals that for a specialized medical task with a limited dataset, leveraging the powerful, frozen features from a large-scale pre-trained model like CLIP is a more robust and effective strategy than end-to-end fine-tuning. The attempts to fine-tune the model likely led to catastrophic forgetting and overfitting, which degraded the quality of the learned representations for both regression and classification. This highlights a critical consideration for applying foundation models in specialized domains: without a sufficiently large fine-tuning dataset, there is a significant risk of harming the model's powerful, pre-trained capabilities.

\section{Methodology}

The architecture of our framework, \ourwork{}, is shown in Figure~\ref{fig:architecture}, which consists of four core components: (1) a CLIP image encoder that extracts semantic features from each viewpoint, (2) a GAT that models inter-pose relationships to produce consistent multi-view predictions, (3) a retrieval module that queries a curated KB using global embeddings to provide representative support, and (4) a context-aware fusion mechanism that adaptively combines GAT and retrieval predictions based on confidence and local embedding density. This design enables generalization under pose variability, class imbalance, and domain shifts.

\textbf{Problem Statement:} Given a set of multi-pose RGB images \(X_i = \{x_{i,1}, x_{i,2}, \dots, x_{i, P_i}\}\) for a subject \(i\), where each \(x_{i,j}\) corresponds to the \( j^{\text{th}} \) available viewpoint (e.g., frontal, lateral, back, selfie), and a scalar age value \(a_i\), the task is to jointly predict:
(i) a binary nutritional status label \(y_i^{\text{class}} \in \{0,1\}\), and 
(ii) a 4-dimensional vector of anthropometric measurements \(y_i^{\text{reg}} \in \mathbb{R}^4\), including Height (Ht.), Weight (Wt.), MUAC, and Head Circumference (HC).

For each subject $i$, we obtain $P$ pose images. Each is embedded via a pretrained CLIP encoder into a 1024D vector. These are concatenated with age to form node features $v_{i,j} \in \mathbb{R}^{1025}$, which are used as nodes in a fully connected undirected graph. This design allows GATs to model inter-pose dependencies while retaining pose-specific signals.

To enhance sensitivity to malnourished cases and support cross-population generalization, we introduce a retrieval-augmented module. A global embedding is computed by averaging the pose embeddings.
Appending age yields a 1025D RAC query, $q_i$, which retrieves top-$k$ similar samples from a FAISS-based KB. Final predictions combine GAT and retrieval outputs using a learned weight. This adaptive fusion supports robustness to class imbalance, pose variability, and domain shift. These modules are described below.
\noindent\textbf{Multi-Pose Embedding Extraction:}
Each image \(x_{i,j}\) is passed through a frozen CLIP encoder (ResNet-50x64 variant)~\cite{radford2021learning}, producing a 1024-dimensional embedding \(e_{i,j} \in \mathbb{R}^{1024}\). To incorporate subject metadata, we append the scalar age \(a_i\) to each embedding, forming enriched pose-level feature vectors:
\(
v_{i,j} = [e_{i,j}; a_i] \in \mathbb{R}^{1025}.
\)
The multi-pose design (top-left block in Figure~\ref{fig:architecture}) offers two advantages: (i) it mitigates the limitations of any single pose by aggregating redundant cues across views, and (ii) it enables generalisation across variable capture conditions (e.g., occlusion or missing poses), critical for deployment in uncontrolled field settings.

\noindent\textbf{Graph Construction and GAT Inference:}
The enriched pose-level embeddings \(\{v_{i,1}, \dots, v_{i,P_i}\}\) are treated as nodes \(V_i\) of a fully connected undirected graph \(G_i = (V_i, E_i)\). Each edge in \(E_i\) links a pair of poses \((v_{i,j}, v_{i,k})\) with \(j \neq k\). This representation enables pairwise message exchange between all poses, allowing the model to reason over inter-pose correlations such as asymmetrical fat loss or posture-related distortions, both of which are often indicative of malnutrition. The resulting graph is processed using a two-layer GAT, where each layer applies multi-head self-attention to selectively aggregate information from neighboring nodes. The final node representations are then globally pooled to obtain a subject-level embedding \(h_i\), which is subsequently passed to task-specific heads for regression and classification.By leveraging cross-pose attention, the GAT module improves robustness to pose imbalance and occlusion while also enhancing interpretability through attention weights computed across views.

\subsection{Knowledge Base Construction}
\label{sec:kb}

To support retrieval-based inference, we construct a curated KB of pose-level embeddings and ground-truth labels from clinically collected data. The KB consists of 1984 multi-view RGB images captured from 248 pediatric subjects.

\noindent\textit{Image Capture Protocol:} Images were acquired using a standard consumer-grade smartphone (OnePlus Nord) equipped with a Sony IMX586 sensor (48MP, f/1.75 aperture, OIS/EIS). Capture guidelines prescribed an approximate subject distance of 165 cm and a camera height of 50 inches, but minor variability was allowed to improve model robustness. Each subject was photographed across eight views: four frontal, one lateral-left, one lateral-right, one posterior, and one frontal selfie.

\noindent\textit{Ground-Truth Labels:} Trained healthcare workers recorded anthropometric measurements, including Height (H), Weight (W), Middle-Upper Arm Circumference (MUAC), and Head Circumference (HC), using standardized clinical protocols. All data entries were logged via a GUI-assisted tool to ensure consistency. The resulting dataset forms the retrieval corpus for nearest-neighbor lookups during inference. Each subject’s global pose-averaged embeddings and labels are indexed using FAISS for efficient similarity search.

\subsection{Retrieval-Augmented Classification}
\label{sec:retrieval_classification}

% To improve sensitivity to the underrepresented malnourished class, we augment the GAT classifier with a FAISS‑based retrieval head. For each subject \(i\), we first compute a global query embedding
% \[
% q_i = \frac{1}{P_i}\sum_{j=1}^{P_i}v_{i,j}\in\mathbb R^{1025},
% \]
To improve sensitivity to the underrepresented malnourished class, we augment the GAT classifier with a FAISS‑based retrieval head. For each subject~\(i\), we first compute a global query embedding \( q_i = \frac{1}{P_i} \sum_{j=1}^{P_i} v_{i,j} \in \mathbb{R}^{1025} \). Then retrieve its top‑\(k\) nearest neighbors from the KB, obtaining cosine distance \(\{d_j\}_{j=1}^k\) and corresponding binary labels \(\{y_j^{\text{kb}}\}_{j=1}^k\). We normalize the cosine distances with a temperature‑scaled softmax,
\[
\tilde w_j \;=\;\frac{\exp(-d_j/\tau)}{\sum_{m=1}^k\exp(d_m/\tau)}\,,
\]
Here, $d_j$ is the cosine distance to the \( j^{\text{th}} \) retrieved neighbor, $k$ is the number of retrieved samples, $\tau$ controls the sharpness of the softmax, and $\tilde{w}_j$ denotes the normalized retrieval weight.
then apply a class‑specific boost
\[
\omega_j \;=\;\begin{cases}
\gamma,&y_j^{\text{kb}}=1\ (\text{malnourished}),\\
1,&y_j^{\text{kb}}=0\ (\text{healthy}),
\end{cases}
\]
Here, $\omega_j$ is a multiplicative class-specific factor that upweights malnourished neighbors by $\gamma$ (when $y^{kb}_j=1$) and leaves healthy neighbors unchanged (when $y^{kb}_j=0$).

The adjusted weights are re‑normalized across all neighbors $m$, 
\(\;w_j\!=\!\tilde w_j\omega_j\big/\sum_{m}\tilde w_m\omega_m,\)
and the retrieval‑based prediction is
\[
y_i^{\text{retrieved}}
=\sum_{j=1}^k w_j\,y_j^{\text{kb}}\,.
\]

Next, we compute an auxiliary context vector
\[
c_i=\bigl[\log\!\tfrac{p_i}{1-p_i},\;\bar d\bigr],
\]
where \(p_i=\sigma(y_i^{\text{GAT}})\) ($\sigma(\cdot)$ = sigmoid) and \(\bar d=\frac1k\sum_jd_j\). A small MLP then predicts a fusion coefficient \(\alpha\in[0,1]\) from \([\,y_i^{\text{GAT}},\,y_i^{\text{retrieved}},\,c_i\,]\), and we form the final logit
\[
\hat y_i^{\text{CLS}}
=\alpha^{\text{CLS}}\;y_i^{\text{GAT}}
+(1-\alpha^{\text{CLS}})\;y_i^{\text{retrieved}}
\]

This adaptive fusion automatically shifts weight toward retrieval when the KB is dense around \(q_i\), and relies on the GAT when neighbors are sparse. 

% All retrieval hyperparameters (\(\gamma\), \(\tau\), \(k\)) were selected via held-out validation on the AV set and shown to be robust to small perturbations (see Supplementary). 

% We extend the retrieval-based mechanism to the regression task to improve estimation of anthropometric measurements in low-data regimes. 

\subsection{Retrieval-Augmented Regression}
Similar to classification, we compute the global query embedding \(q_i\) for subject \(i\) by averaging their pose-level embeddings and use FAISS to retrieve the top-\(k\) most similar subjects from the KB. Each retrieved sample contributes a ground-truth vector \(y_j^{\text{reg}} \in \mathbb{R}^4\), containing four measurements: Ht, Wt, MUAC, and HC.

The cosine distances \(\{d_1, \dots, d_k\}\) are passed through a softmax to produce normalized weights \(\{w_1, \dots, w_k\}\), for retrieval-based regression:
\[
y_i^{\text{retrieved}} = \sum_{j=1}^{k} w_j \cdot y_j^{\text{reg}}.
\]

This estimate is fused with the GAT-based regression output \(y_i^{\text{GAT}}\) using a learnable scalar fusion coefficient \(\alpha \in [0,1]\), yielding the final prediction:
\[
\hat{y}_i^{\text{reg}} = \alpha^{\text{reg}} \cdot y_i^{\text{GAT}} + (1 - \alpha^{\text{reg}}) \cdot y_i^{\text{retrieved}}
\]

% By leveraging both structured inter-pose reasoning and contextual knowledge (KB), this approach enables accurate regression even in sparse or imbalanced training settings.
All retrieval hyperparameters ($k$, $\tau_{\text{class}}$, $\gamma$, $\tau_{\text{reg}}$) were tuned on the AnthroVision validation set. 
Here, $k$ is the number of retrieved neighbors, $\tau_{\text{class}}$ and $\tau_{\text{reg}}$ control similarity weighting for classification and regression, and $\gamma$ upweights malnourished exemplars. 
Fusion coefficients $\alpha^{\text{CLS}}$ and $\alpha^{\text{reg}}$ are learned during training.

\section{Experiments}
% \subsection{Experimental settings}

% \textbf{Datasets:} We evaluate our models on three cross-continental multipose image datasets encompassing children and adults, collected from multiple sources. For each subject, images are captured from multiple anatomical viewpoints (e.g., frontal, lateral, back) and paired with demographic and anthropometric measurements where available. 

\textbf{Datasets:} We evaluate on three cross-continental multi-pose datasets comprising child and adult subjects. Each sample includes images from various anatomical views (e.g., frontal, lateral, back) with corresponding demographic and anthropometric data.

\begin{itemize}
    \item AnthroVision~\cite{khan2024domainadapt}: Our primary dataset with $2,141$ children from clinical and community settings, containing multipose images and detailed labels (gender, Ht (cm), Wt (g), HC (cm), waistline (cm), MUAC (cm), and age (months)). Binary malnutrition labels are derived from WHO z-score thresholds~\cite{WHOChildgrowthstandards}. It serves as our core dataset for training, validation, and ablation (Being the largest available dataset with malnutrition and anthropometric labels).
    
    % \item AnthroVision~\cite{khan2024domainadapt}: This is our primary dataset consisting of $2,141$ children from both clinical and community settings, with comprehensive multipose images and detailed ground-truth labels: gender, Ht (cm), Wt (g), HC (cm), waistline (cm), MUAC (cm), and age (months). Each subject has a binary malnutrition label determined by standard anthropometric z-score~\cite{WHOChildgrowthstandards} thresholds. This data has the largest sample size amongst all available malnourishment and anthropometric measurement data, and so we use this dataset for core training, validation, and ablation studies.
    \item The Age-Restricted Anonymized (ARAN)~\cite{MohammedKhan2025ARAN}: Contains 512 children (16–98 months) from two hospitals (H1: 404, H2: 108), with four face-anonymized views and measurements for Ht, Wt, waist, and HC. 
    \item CampusPose: An in-house college-aged dataset (80 subjects) with multipose images and measurements for Ht, Wt, MUAC, HC and WC, and age. Malnutrition labels are based on BMI thresholds.
\end{itemize}

Ethical clearance for this data collection was obtained from the Institutional Ethics Committee (IEC) IIT-AIIMS Jodhpur, India.

        % \item A2: Another adult dataset from~\cite{liu2018single} (31 subjects), with triplicate MUAC, upper-arm, and head circumference readings. Labels for malnourishment derived post-hoc using BMI thresholds~\cite{WHOChildgrowthstandards}.
    % \item MalDB dataset : An in-house collection of 3,090 online-obtained images labeled by medical professionals as malnourished. These serve as an external test-only set, used solely for assessing generalisation on in-the-wild samples for binary malnourishment classification.
    % \item MalKB: For retrieval-augmented evaluation, we construct KBs of inhouse collected 248 representative samples comprising of 1984 images. 

% \begin{table}[!tt]
% \small
% \centering
% \setlength{\tabcolsep}{2pt} % Adjust as needed for column fit
% \begin{tabular}{|p{2.7cm}|p{1.3cm}|p{1.3cm}|p{1.5cm}|p{0.7cm}|}
% \hline
% \textbf{Dataset} & \textbf{N / M / F} & \textbf{N Images} & \textbf{Age Range} & \textbf{Poses} \\
% \hline
% AnthroVision (AV)      &          &          &          &          \\
% ARAN    &          &          &          &          \\
% Adult A1      &          &          &          &          \\
% %  
% % MalDB     &          &          &          &          \\
% MalKB &          &          &          &          \\
% \hline
% \end{tabular}
% \caption{Summary of datasets used in this study. N/M/F: Number of subjects, males, and females.}
% \label{tab:dataset-summary}
% \end{table}

\begin{figure*}[t]
\centering

\begin{minipage}[t]{0.32\linewidth}
    \centering
    \includegraphics[width=\linewidth]{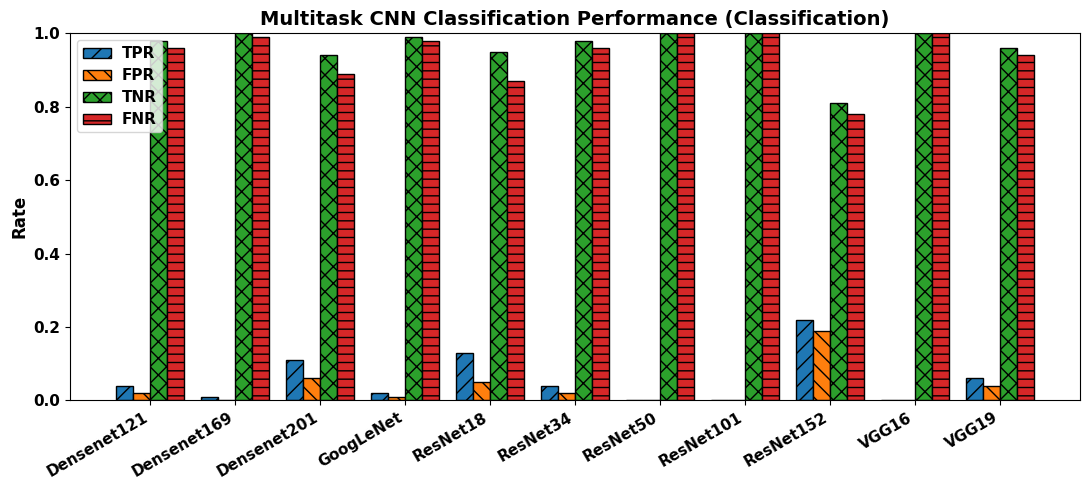}
    \caption*{(a) CNN classification error rates.}
    \label{fig:cnn_class_bar}
\end{minipage}
\hfill
\begin{minipage}[t]{0.35\linewidth}
    \centering
    \includegraphics[width=\linewidth]{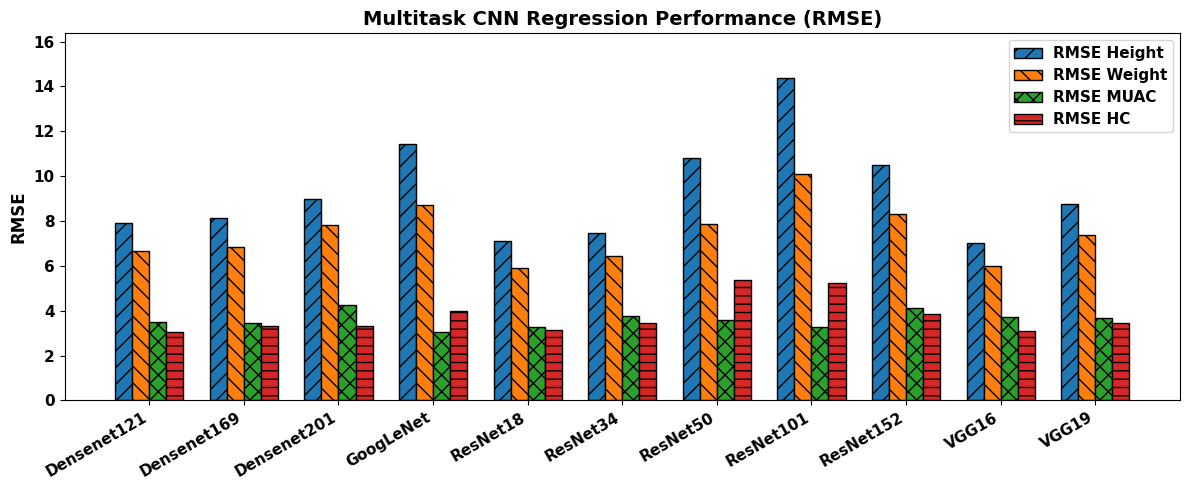}
    \caption*{(b) CNN regression RMSE.}
    \label{fig:cnn_reg_bar}
\end{minipage}
\hfill
\begin{minipage}[t]{0.32\linewidth}
    \centering
    \includegraphics[width=0.75\linewidth]{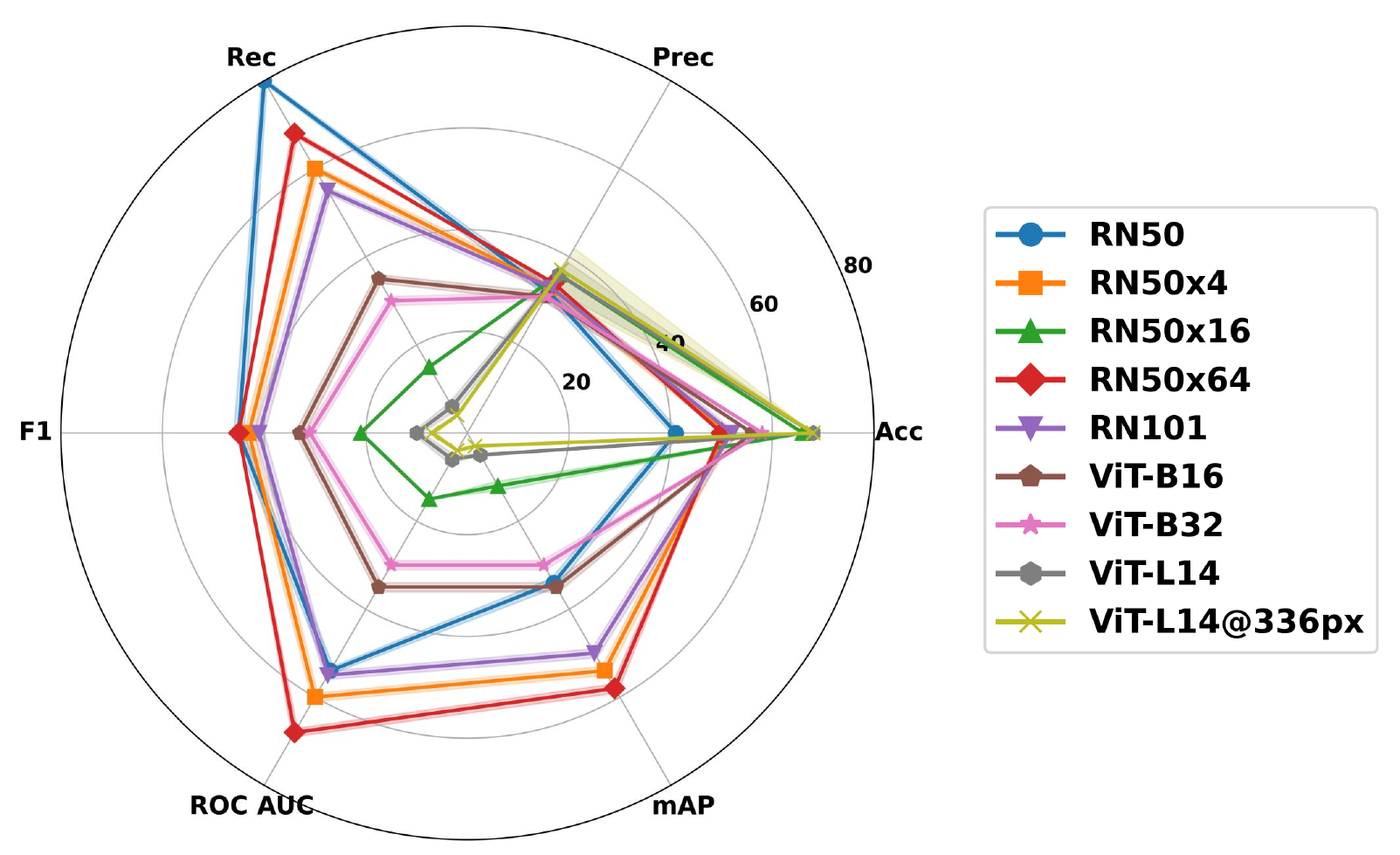}
    \caption*{(c) CLIP encoder comparison.}
    \label{fig:radar_clip}
\end{minipage}

% \vspace{2mm}
\caption{Baseline model performance across (a) classification errors, (b) regression RMSE, and (c) encoder variants.}
\label{fig:baseline_trio}
\end{figure*}

\subsection{Implementation Details}

\noindent\textbf{Training Objectives:} We train the network end-to-end with a joint loss \(\mathcal{L} = \mathcal{L}_{\text{class}} + \mathcal{L}_{\text{reg}}\) (binary cross-entropy with logits loss (BCElogitloss) and mean squared error). Implementations are in PyTorch and PyTorch Geometric with 4-fold cross-validation with random seed 42. Workstation for experiments runs on Ubuntu 20.04 LTS with four NVIDIA A100 GPUs (40GB each), 128GB RAM, Python 3.10, PyTorch 2.1.0, CUDA 11.8, and cuDNN 8.4. The model begins with a frozen CLIP (RN50x64) backbone and processes 1025-dimensional pose embeddings via a 2-layer GAT (8 heads, dropout 0.1). Training uses a batch size of 8 for 50 epochs (early stop) per fold with the Adam optimizer (\(1\times10^{-3}\) learning rate). 

\noindent\textbf{Evaluation Metrics:} Classification metrics include accuracy, precision, recall, F1 score, ROC AUC, and mAP, while regression is evaluated with RMSE and MAE. We also tried maximizing F1 and balanced accuracy; Youden’s index gave the best trade-off between sensitivity and specificity.

\noindent\textbf{Component Benchmarking:} We benchmark different components of \ourwork{} in the following manner:

\noindent\textbf{1. CNN-Based Predictors:} We evaluate a range of standard convolutional backbones for joint classification and regression. Architectures include ResNet (18/34/50/101/152), DenseNet (121/169/201), GoogLeNet, and VGG (16/19), all pretrained on ImageNet. Each pose image is processed independently. 

\noindent\textbf{2. CLIP Encoder Variants:}
We benchmark multiple CLIP variants (RN50, RN50x4, RN50x16, RN50x64, RN101, ViT-B16, ViT-B32, ViT-L14, ViT-L14@336px) for pose-wise classification. Each pose image is encoded as an individual input for CLIP Zero Shot Evaluation.

% using a frozen CLIP model and passed to a linear classifier. 

\noindent\textbf{3. Pretrained vs. Fine-Tuned CLIP in End-to-End Pipeline:} We test whether fine-tuning a selected CLIP (RN50×64) encoder on our dataset improves performance. Both image and text encoders were fine-tuned using prompts like ``\textit{\{pose\} photo of a \{malnourished/healthy} child\},'' with the rest of the NutriScreener pipeline (GAT, retriever, fusion) unchanged. We then compared the fine-tuned and frozen variants within the full framework.

% \paragraph{4. Retrieval Alternatives:}
% We compare our retrieval-augmented classification (RAC) module with distance-based baselines, including Euclidean and Mahalanobis retrieval. 

% \textbf{todo: transformers, BERT, ViT}
% \textbf{todo: pretrained/scratch/fine-tuned cnn}
% \textbf{todo: CNN Uncertainity/Gradnorm}

\subsubsection{Baseline Comparisons} We benchmark \ourwork{} against existing methods and ablations under three settings:

\noindent\textbf{1. Baseline Models for Malnutrition Detection:} We compare different versions of \ourwork{} with DomainAdapt~\cite{khan2024domainadapt}. Other versions include: 

\begin{itemize}
    \item \textbf{DomainAdapt}~\cite{khan2024domainadapt}: Prior benchmark multitask model that combines classification and regression.
    \item \textbf{\ourwork{} (BCE)}: Uses weighted binary cross-entropy loss on the GAT output to up-weight malnourished samples.
    \item \textbf{\ourwork{} (Focal)}: Replaces BCE with focal loss to emphasize hard-to-classify minority-class samples.
    \item \textbf{\ourwork{} (Context)}: Adds calibrated log-odds and local density estimates as auxiliary inputs to the fusion module. Trained with weighted BCE.
    \item \textbf{\ourwork{} (Weighted)}: Applies temperature-scaled retrieval weighting with minority-class boosting, combined with context and weighted BCE.
\end{itemize}

\noindent\textbf{2. Cohort Generalization:} To assess robustness across settings, we split AnthroVision into clinical and community cohorts and report AUC and mAP for both DomainAdapt and our strongest variant from previous baselining.

\noindent\textbf{3. Cross Domain Anthropometric Comparison:} We compare against SOTA estimation networks from related domains such as Altinigne et al. \cite{altinigne2020height} (IMDB, full-body) and Dantcheva et al. \cite{dantcheva2018show} (VIP, face).

% While prior work on malnutrition screening is limited, we compare our regression performance (MAE for height and weight) against state-of-the-art anthropometric estimation models:
% \begin{itemize}
%     \item Altinigne et al.~\cite{altinigne2020height}: Report MAEs using full-body images from the IMDB dataset.
%     \item Dantcheva et al.~\cite{dantcheva2018show}: Estimate anthropometric attributes from facial images in the VIP Attribute dataset.
% \end{itemize}

\noindent\textbf{Cross-Dataset Analysis} We evaluate the generalization ability of \ourwork{} by varying the retrieval KB while keeping the training set fixed to AnthroVision with the largest sample size. Different KBs, such as MalKB, ARAN subsets (H1, H2), and CampusPose dataset, are used at inference to assess their impact across test cohorts. 

% The comparative visualisation of all involved datasets embeddings is shown in Figure \ref{fig:tsne}.

\subsection{Overall Results}

\noindent\textbf{CNN and VLM Baselines.} 
Figures~\ref{fig:baseline_trio} (a) and (b) show that standard CNN-based multitask models (ResNet, DenseNet, VGG, GoogLeNet) struggle with malnutrition classification, exhibiting high anthropometric RMSEs and low recall due to class imbalance and lack of pose-aware structure. To explore stronger backbones, we evaluated vision-language model (VLM) encoders, comparing multiple CLIP variants on per-pose image classification (Figure~\ref{fig:baseline_trio} (c)). Among them, RN50$\times$64 achieved the highest ROC AUC (68\%) and mAP (58\%), with balanced recall and precision, outperforming smaller variants (e.g., RN50 overpredicts with recall: 80\%, precision: 32\%) and larger ones (e.g., RN50$\times$16 underfits with recall: 15\%). This supports RN50$\times$64 as the most robust and balanced encoder under class imbalance.

% Metrics: Accuracy (Acc), Precision (Prec), Recall (Rec), F1-score (F1), Area Under Curve (AUC), mean Average Precision (mAP), and RMSE for anthropometric estimates (Height, Weight, MUAC, Head Circumference).
\begin{table*}[t]
\centering
\caption{Performance comparison of baseline models and the proposed NutriScreener (Weighted) model on the AnthroVision dataset. DA = DomainAdapt, C+GNN = CLIP+GNN, Ret-only = Retrieval-only, NS-W = NutriScreener (Weighted). Regression metric is RMSE in ``\textit{cm}'' (H, MUAC, HC) and ``\textit{kg}'' (W)}

\fontsize{9pt}{11pt}\selectfont
\setlength{\tabcolsep}{2pt}

\begin{tabular}{|l|c|c|c|c|c|c|c|c|c|c|}
\hline
\rowcolor{gray!20}
\textbf{Model} & \textbf{Acc ↑} & \textbf{Prec ↑} & \textbf{Rec ↑} & \textbf{F1 ↑} & \textbf{AUC ↑} & \textbf{mAP ↑} & \textbf{H ↓} & \textbf{W ↓} & \textbf{MUAC ↓} & \textbf{HC ↓} \\ \hline
DA & 0.68 ± 0.01 & 0.63 ± 0.01 & 0.67 ± 0.04 & 0.64 ± 0.03 & 0.55 ± 0.02 & 0.35 ± 0.02 & 22.00 ± 1.07 & 12.40 ± 0.32 & 3.55 ± 0.24 & 5.05 ± 0.18 \\
C+GNN & \textbf{0.76 ± 0.03} & \textbf{0.66 ± 0.04} & 0.54 ± 0.07 & 0.59 ± 0.04 & \textbf{0.82 ± 0.03} & \textbf{0.66 ± 0.03} & 7.37 ± 0.55 & 5.82 ± 0.30 & 3.80 ± 0.28 & 5.23 ± 0.22 \\
Ret-only & 0.53 ± 0.04 & 0.36 ± 0.06 & 0.66 ± 0.05 & 0.45 ± 0.03 & 0.61 ± 0.07 & 0.38 ± 0.02 & 9.48 ± 0.37 & 7.89 ± 0.63 & 3.12 ± 0.47 & 2.76 ± 0.28 \\
\rowcolor{gray!20}
NS-W & 0.74 ± 0.04 & 0.56 ± 0.01 & \textbf{0.79 ± 0.02} & \textbf{0.66 ± 0.01} & \textbf{0.82 ± 0.01} & 0.65 ± 0.02 & \textbf{6.38 ± 0.46} & \textbf{5.32 ± 0.56} & \textbf{2.80 ± 1.47} & \textbf{2.97 ± 1.54} \\
\hline
\end{tabular}
\label{tab:unified_baselines}
\end{table*}

\begin{table*}[t]
\centering
\small
\caption{Ablation study of NutriScreener variants. Regression metric is RMSE in ``\textit{cm}'' (H, MUAC, HC) and ``\textit{kg}'' (W)}
\fontsize{9pt}{11pt}\selectfont
\setlength{\tabcolsep}{2pt}
\begin{tabular}
{|l|c|c|c|c|c|c|c|c|c|c|}
\hline
\rowcolor{gray!20}
\textbf{Variant} & \textbf{Acc ↑} & \textbf{Prec ↑} & \textbf{Rec ↑} & \textbf{F1 ↑} & \textbf{AUC ↑} & \textbf{mAP ↑} & \textbf{H ↓} & \textbf{W ↓} & \textbf{MUAC ↓} & \textbf{HC ↓} \\ \hline
BCE & 0.66 ± 0.07 & 0.47 ± 0.04 & \textbf{0.81 ± 0.12} & 0.59 ± 0.04 & 0.78 ± 0.04 & 0.60 ± 0.03 & 10.93 ± 0.48 & 8.33 ± 0.26 & 3.88 ± 0.22 & 3.15 ± 0.19 \\
Focal & 0.62 ± 0.07 & 0.42 ± 0.02 & 0.73 ± 0.07 & 0.53 ± 0.01 & 0.73 ± 0.04 & 0.52 ± 0.03 & 10.82 ± 0.45 & 8.46 ± 0.28 & 4.28 ± 0.24 & 3.33 ± 0.20 \\
Context & 0.73 ± 0.04 & 0.54 ± 0.02 & 0.65 ± 0.01 & 0.59 ± 0.01 & 0.78 ± 0.04 & 0.58 ± 0.03 & 10.82 ± 0.45 & 8.46 ± 0.28 & 4.28 ± 0.24 & 3.33 ± 0.20 \\
\rowcolor{gray!20}
NS-W & \textbf{0.74 ± 0.04} & \textbf{0.56 ± 0.01} & 0.79 ± 0.02 & \textbf{0.66 ± 0.01} & \textbf{0.82 ± 0.01} & \textbf{0.65 ± 0.02} & \textbf{6.38 ± 0.46} & \textbf{5.32 ± 0.56} & \textbf{2.80 ± 1.47} & \textbf{2.97 ± 1.54} \\ \hline
\end{tabular}
\label{tab:main_ablation_transposed}
\end{table*}

\noindent\textbf{Impact of Fine-Tuning.} 
To test whether domain-specific adaptation improves generalization, we fine-tuned both the image and text encoders of RN50$\times$64 using prompts of the form: \textit{\{pose\} photo of a \{malnourished/healthy\} child}. This encoder was plugged into the downstream NutriScreener pipeline (GNN, retrieval, fusion). The frozen pretrained encoder outperformed the fine-tuned version across all metrics, malnutrition recall (79\% vs. 38\%), Ht RMSE (6.38 vs. 8.87 cm), ROC-AUC (0.82 vs. 0.72), and mAP (0.65 vs. 0.54). These findings align with prior work cautioning against fine-tuning foundation models in low-resource settings due to representational collapse~\cite{kumar2022finetuning, gong2023arf}. These findings suggest that the pretrained CLIP features, learned from large-scale image-text pairs, encode semantically relevant anthropometric cues. Freezing the encoder preserves these representations and avoids overfitting to dataset-specific correlations, leading to better generalization under data scarcity.

% \begin{figure}[t]
%     \centering
%     % Top barplot: classification metrics
%     \begin{subfigure}{\linewidth}
%         \centering
%         \includegraphics[width=240pt]{AnonymousSubmission/LaTeX/images/cnn classification.png}
%         \caption*{Classification error rates (TPR, FPR, TNR, FNR) for malnourished class.}
%         \label{fig:cnn_class_bar}
%     \end{subfigure}

%     \vspace{2mm}

%     % Bottom barplot: regression metrics
%     \begin{subfigure}{\linewidth}
%         \centering
%         \includegraphics[width=240pt]{AnonymousSubmission/LaTeX/images/cnn regression.png}
%         \caption*{Regression performance (RMSE) for Height, Weight, MUAC, and HC.}
%         \label{fig:cnn_reg_bar}
%     \end{subfigure}

%     \caption{Performance of baseline CNN models across (a) classification and (b) regression tasks.}
%     \label{fig:cnn_error_bars}
% \end{figure}

% \begin{figure}[!t]
%     \centering
%     \includegraphics[width=0.7\linewidth]{AnonymousSubmission/LaTeX/images/my_radar_chart.pdf}
%     \caption{Comparison of different CLIP encoders (ResNet-~\cite{radford2021learning} and ViT-based~\cite{dosovitskiy2020image}).}
%     \label{fig:radar_clip}
% \end{figure} 

% community and clinical
\begin{table}[t]
\centering
\caption{Comparison of community and clinical performance with baseline}
\label{tab:split_performance}
\fontsize{9pt}{11pt}\selectfont
\begin{tabular}{lcccc}
\hline
 & \multicolumn{2}{c}{AUC} & \multicolumn{2}{c}{mAP} \\
\cline{2-5}
Model & Comm & Clin & Comm & Clin \\
\hline
DomainAdapt & 0.64 & 0.51 & 0.40 & 0.35 \\
\rowcolor{gray!20}
\ourwork{} (Weighted) & 0.78 & 0.74 & 0.53 & 0.58 \\
\hline
\end{tabular}
\end{table}

\begin{table}[t]
\centering
\caption{Existing work reporting mean absolute error (MAE) for Ht and Wt estimation from images.}
\label{tab:existingMAE}
\fontsize{9pt}{11pt}\selectfont
\begin{tabular}{l l c c}
\hline
Dataset & Image Type & MAE (H) & MAE (W) \\
\hline
IMDB Dataset & Full-Body & 6.13 cm & 9.80 kg \\
VIP Attribute Dataset & Face & 8.20 cm & 8.51 kg \\
\rowcolor{gray!20}
AnthroVision & Multi-Pose & 4.69 cm & 3.78 kg \\
\hline
\end{tabular}
\end{table}

\begin{table*}[t]
\centering
\caption{Impact of KB dataset choice on classification and regression performance. Each test set is evaluated using different KBs. ``-'' means there is no available ground truth in that dataset. C: Classification, R: Regression. \textbf{Note:} Highly out-of-distribution KBs (CampusPose cohort) result in identical performance to No Retrieval pertaining to the fusion mechanism. CP: CampusPose. Regression metric is RMSE in ``\textit{cm}'' (H, MUAC, HC) and ``\textit{kg}'' (W)}
\label{tab:cross_dataset}
\small
\setlength{\tabcolsep}{3pt}
\renewcommand{\arraystretch}{1.1}
\begin{tabular}{p{0.7cm} p{0.7cm} p{1.5cm} c c c c c c c c c}
\toprule
\textbf{Train} & \textbf{Test} & \textbf{Labels} & \textbf{KB} & \textbf{Rec} & \textbf{F1} & \textbf{AUC} & \textbf{mAP} & \textbf{H} & \textbf{W} & \textbf{MUAC} & \textbf{HC} \\
\midrule
\multirow{5}{*}{AV} & \multirow{5}{*}{AV} & \multirow{5}{*}{C+R} 
& NoRet   & 0.54±0.15 & 0.59±0.04 & 0.82±0.03 & 0.66±0.05 & 7.37±0.55 & 5.82±0.30 & 3.80±1.40 & 5.23±1.60 \\
&&& H2    & 0.54±0.15 & 0.59±0.04 & 0.82±0.03 & 0.66±0.05 & 7.37±0.55 & 5.82±0.30 & 3.80±1.40 & 5.23±1.60 \\
&&& MalKB+H2    & 0.73±0.14 & 0.60±0.04 & 0.80±0.02 & 0.62±0.03 & 6.53±0.50 & 5.96±0.26 & 3.32±1.44 & 2.93±1.56 \\
&&& CP   & 0.66±0.18 & 0.58±0.03 & 0.80±0.02 & 0.62±0.04 & 7.37±0.55 & 5.82±0.30 & 3.80±1.40 & 5.23±1.60 \\
\rowcolor{gray!20}
&&& MalKB   & 0.79±0.02 & 0.66±0.01 & 0.82±0.01 & 0.65±0.02 & 6.38±0.46 & 5.32±0.56 & 2.80±1.47 & 2.97±1.54 \\
\midrule
\multirow{3}{*}{AV} & \multirow{3}{*}{H1} & \multirow{3}{*}{R: H, W, HC}
& NoRet   & - & - & - & - & 7.87±0.30 & 5.91±0.47 & - & 4.59±1.98 \\
&&& MalKB   & - & - & - & - & 7.87±0.30 & 5.91±0.47 & - & 4.59±1.98 \\
&&& CP   & - & - & - & - & 7.87±0.30 & 5.91±0.47 & - & 4.59±1.98 \\
\rowcolor{gray!20}
&&& H2   & - & - & - & - & 7.21±0.02 & 5.67±0.38 & - & 4.20±0.18 \\
\midrule
\multirow{4}{*}{AV} & \multirow{4}{*}{CP} & \multirow{4}{*}{C+R}
& NoRet   & 0.75±0.01 & 0.63±0.04 & 0.64±0.04 & 0.69±0.02 & 23.98±0.27 & 17.36±0.16 & 9.46±0.28 & 3.98±0.31 \\
&&& MalKB   & 0.75±0.01 & 0.63±0.04 & 0.64±0.04 & 0.69±0.02 & 23.98±0.27 & 17.36±0.16 & 9.46±0.28 & 3.98±0.31 \\
&&& H2    & 0.72±0.04 & 0.65±0.01 & 0.65±0.02 & 0.69±0.02 & 23.98±0.27 & 17.36±0.16 & 9.46±0.28 & 3.98±0.31 \\
&&& MalKB+H2    & 0.75±0.03 & 0.63±0.05 & 0.64±0.02 & 0.69±0.03 & 23.98±0.27 & 17.36±0.16 & 9.46±0.28 & 3.98±0.31 \\
\bottomrule
\end{tabular}
\end{table*}

\noindent\textbf{Baseline Comparison with SOTA and Variant Progression.} Tables~\ref{tab:unified_baselines} and ~\ref{tab:main_ablation_transposed} present classification and regression performance across prior baselines and successive NutriScreener variants. 95\% confidence; Friedman and Wilcoxon tests for inter-model and pairwise comparisons reported in \textit{Appendix}. The literature baseline, \emph{DomainAdapt}~\cite{khan2024domainadapt}, achieves moderate classification (Recall: 0.67, F1: 0.64, AUC: 0.55) but suffers from poor regression accuracy (e.g., Ht RMSE: 22.0 cm), highlighting limited generalization under class imbalance. A stronger non-retrieval baseline, \emph{Baseline GNN}, integrates multi-pose CLIP (RN50x64) embeddings and age metadata via a GAT, yielding significant gains in AUC (0.82) and regression (Ht: 7.37 cm, Wt: 5.82 kg), though recall remains low (0.54), indicating insufficient minority-class modeling. Introducing retrieval augmentation, \emph{NutriScreener (BCE)} fuses FAISS-based neighbors with class-weighted BCE loss, improving recall (0.81) but degrading precision (0.47) and regression (Ht RMSE: 10.93 cm), suggesting over-sensitivity to noisy retrieved samples. To mitigate this, \emph{NutriScreener (Focal)} replaces BCE with focal loss, which maintains high recall (0.76) but lowers F1 (0.53) and inflates MUAC error (4.28 cm), indicating over-penalization of easy samples. To better filter retrieved support, \emph{NutriScreener (Context)} introduces calibrated log-odds and local density as auxiliary features during fusion, leading to balanced precision (0.54), improved F1 (0.59), and stable AUC (0.78), with minimal regression degradation. Finally, \emph{NutriScreener (Weighted)} applies temperature-scaled, class-boosted weighting over retrieved neighbors, resulting in the best overall metrics: Recall (0.79), F1 (0.66), AUC (0.82), and the lowest RMSEs across all anthropometric variables (Ht: 6.38 cm, Wt: 5.32 kg, MUAC: 2.80 cm, HC: 2.97 cm). 

\noindent\textbf{Cohort Generalization.} Table \ref{tab:split_performance} highlights the generalization strength of NutriScreener (Weighted) across both community and clinical settings, achieving AUC scores of 0.78 and 0.74, respectively, significantly outperforming DomainAdapt. The consistent mAP improvement confirms its robustness to population and context shifts, a critical requirement for real-world malnutrition screening. 

\noindent\textbf{Cross Domain Anthropometric Comparison.} Table \ref{tab:existingMAE} presents results across different domains for the shared task of image-based Ht and Wt estimation. NutriScreener outperforms prior methods from adult subject and controlled imaging domains, demonstrating superior generalization and robustness in pediatric, real-world settings.

\noindent\textbf{Cross-Dataset Analysis.} Table~\ref{tab:cross_dataset} shows that the choice of retrieval KB significantly impacts classification and regression outcomes. On the AV test set, using NoRet, CampusPose, or H2 KBs results in lower recall (0.54–0.66) and moderate F1 (0.58–0.59), indicating poor malnutrition sensitivity. In contrast, MalKB achieves superior recall (0.79), F1 (0.66), AUC (0.82), and the lowest RMSEs across anthropometric measures (Ht: 6.37 cm, Wt: 5.32 kg, MUAC: 2.81 cm, HC: 2.97 cm). Even partial KB augmentation (MalKB+H2) improves recall (0.73), highlighting generalization from minor domain expansion. In the ARAN-H1 setting (regression-only), most KBs perform similarly (Ht RMSE: 7.87 cm), but H2 achieves better RMSEs (Ht: 7.21 cm, HC: 4.20 cm), indicating value in intra-cohort matching. For the CampusPose cohort, all KBs yield nearly identical results (recall: 0.72–0.75, F1: 0.60–0.63, AUC: $\sim$0.64, Ht RMSE: 23.98 cm), reflecting domain saturation and the limited utility of retrieval under large population shifts.

\section{User Study: Clinician Feedback}

To evaluate \ourwork{}'s clinical utility, we conducted a user study with 12 medical professionals (mean experience: 9.5 years), including pediatricians and general practitioners. After a brief demo, clinicians received a standalone version of the toolkit, containing the trained model and knowledge base as embedded, non-reversible structures, and applied it to an average of 15 pediatric cases per doctor during regular clinical workflow. Feedback via Likert-scale indicated strong acceptance: clinical consistency (4.3/5), efficiency (4.6/5), trustworthiness (4.4/5), and deployment readiness (4.1/5). Participants especially valued the tool as an objective ``second opinion,'' with one noting it successfully flagged a visually ambiguous malnourishment case. Open-ended responses suggested key improvements, including the mention of uncertainty and visual cues to highlight key visual areas. Overall, clinicians found \ourwork{} reliable, efficient, and suitable for real-world deployment, particularly by community health workers in low-resource settings.

\section{Conclusion and Future Work}

Evaluated on cross continent datasets (AnthroVision, ARAN, and CampusPose), \ourwork{} establishes a new benchmark for child malnutrition screening, addressing both algorithmic and deployment-level challenges in low-resource contexts. By integrating class-boosted context-aware retrieval augmentation with multi-pose graph attention over CLIP features, \ourwork{} achieves strong sensitivity (Recall: 0.79), generalization (AUC: 0.82), and low anthropometric prediction errors, outperforming CNN and domain-adaptive baselines. Cross-dataset analysis shows that even a small, demographically aligned knowledge base can yield up to 25\% recall gain and 3.5 cm RMSE reduction, highlighting the framework's adaptability to new populations. Clinician validation (trust: 4.4/5) confirms the system’s accuracy and readiness for low-resource deployment.

Beyond technical gains, \ourwork{} enables scalable, low-cost screening from routine images, reducing manual effort, and as an assistive tool, it supports early detection of at-risk children. It empowers frontline workers and facilitates timely care. Future work will expand the knowledge base diversity and add interpretability and uncertainty for equitable deployment.
\section{Ethics, Safety, and Responsible Deployment}

This study was conducted under institutional ethics approvals from IIT and AIIMS Jodhpur, India. Informed consent was obtained from all participants' parents or guardians. NutriScreener Toolkit follows privacy-by-design principles. It operates on non-reversible CLIP embeddings from which original images cannot be reconstructed, retains no personally identifiable information beyond age and anthropometrics, and stores all data on encrypted, access-controlled infrastructure. To prevent misuse, including unauthorized use for body-shaming, aesthetic judgments, or non-clinical surveillance, NutriScreener will be distributed under a restricted research license that permits only educational and research applications. With a recall of 79\%, NutriScreener prioritizes sensitivity to minimize false negatives and is intended as a research-oriented screening aid rather than a standalone diagnostic tool. The knowledge base (n = 248) comprises consented subjects whose data will be deleted upon their request. Cross-population validation demonstrates consistent performance when demographically matched knowledge bases are used, supporting equitable and generalisable performance.

\section*{Acknowledgments}
M.~Khan is partly supported through the Prime Minister’s Research Fellowship (PMRF), Government of India. This research is also partially supported through Srijan: Center for Excellence in GenAI. 
The authors thank the clinicians and volunteers from AIIMS and IIT Jodhpur for their participation in data collection and field validation.

\bibliography{aaai2026}

\bigskip\bigskip
\begin{center}
\Large\bfseries Supplementary Material
\end{center}
\medskip

\section{Overview}
 Our primary contribution lies in the novel integration of  retrieval-augmented learning, multi-view GNN architectures, and vision-language encoders into a unified pipeline for malnutrition assessment from multi-pose images. The core value of this paper is in the design, implementation, and empirical validation of an end-to-end framework that achieves robust, generalizable performance in a challenging, low-resource clinical setting. We demonstrate that carefully engineered fusion and retrieval mechanisms can bridge the gap between data-driven learning and real-world health deployment. This focus is consistent with the intended impact of our work: enabling practical, scalable malnutrition screening in resource-limited environments.
\section{Architectural and Component Ablations}
\subsection{Pose and Architecture Ablations}
The ablation in \textbf{Table~\ref{tab:supplementary_ablation} (B)} reveals distinct contributions of each pose node in the GNN. Removing the Selfie node slightly reduces recall despite higher accuracy, indicating limited but minority-class-relevant cues. Excluding the Back node yields high recall but poor accuracy, suggesting overprediction of malnourishment. Dropping Frontal or Lateral nodes sharply degrades F1, highlighting their importance for balanced classification. Notably, removing the Lateral node worsens regression more than Frontal, underscoring its importance in anthropometric estimation. We next evaluate architectural and metric-level ablations. \textbf{Classification (Table~\ref{tab:supplementary_ablation} (A):} The 2L-8H-0.1D (\ourwork{} Weighted) setup achieves optimal recall (0.79) and F1 (0.66), though with slightly lower accuracy (0.74) than 2L-4H-0.1D (Acc: 0.76, F1: 0.56). Minimal configurations (2L-2H-0.1D) completely fail to detect malnourished samples (recall: 0), while higher dropout (2L-4H-0.3D) improves recall (0.80) at the cost of specificity. Distance metrics also influence outcomes: Euclidean performs comparably to cosine, but Mahalanobis yields high TPR (0.98) with extremely poor accuracy (0.36) and general reliability. \textbf{Regression (Table~\ref{tab:supplementary_ablation} (A):} 2L-8H-0.1D also achieves the lowest RMSE and MAE across tasks, confirming its ability to leverage multi-pose variance. Both underparameterized (2H) and overparameterized (4L-4H) setups degrade performance. Overall, \textbf{multi-head attention is crucial} for minority-class detection and precise regression, while \textbf{metric choice can skew performance}, with Mahalanobis yielding unreliable results despite high sensitivity.

\begin{table*}[t]
\centering
\small
\caption{Supplementary ablation of retrieval architecture configurations (XL–YH–ZD = Layers–Heads–Dropout) and pose-specific node removal in GNN. Regression targets have the metric RMSE}
\begin{tabular}{|l|c|c|c|c|c|c|c|c|c|c|}
\hline
\rowcolor{gray!20}
\textbf{Variant} & \textbf{Acc} & \textbf{Prec} & \textbf{Rec} & \textbf{F1} & \textbf{AUC} & \textbf{mAP} & \textbf{H} & \textbf{W} & \textbf{MUAC} & \textbf{HC} \\
\hline
\multicolumn{11}{|c|}{\textbf{A. Retrieval architecture ablation (XL–YH–ZD = Layers–Heads–Dropout)}} \\ \hline
\rowcolor{gray!20}
2L-8H-0.1D & 0.74 & 0.56 & 0.79 & \textbf{0.66} & \textbf{0.82} & 0.65 & \textbf{6.38} & \textbf{5.32} & \textbf{2.80}& \textbf{2.97} \\
2L-2H-0.1D & 0.68 & 0.00 & 0.00 & 0.00 & 0.62 & 0.44 & 10.46 & 8.14 & 3.50 & 5.38 \\
2L-4H-0.1D & \textbf{0.76} & 0.60 & 0.53 & 0.56 & 0.82 & 0.57 & 10.92 & 8.66 & 5.98 & 2.18 \\
2L-4H-0.3D & 0.63 & 0.45 & 0.80 & 0.58 & 0.75 & 0.59 & 10.44 & 8.14 & 3.50 & 5.38 \\
3L-4H-0.1D & 0.72 & \textbf{0.71} & 0.19 & 0.30 & 0.79 & 0.60 & 6.78 & 5.87 & 2.87 & 5.29 \\
4L-4H-0.1D & 0.34 & 0.32 & 0.96 & 0.48 & 0.58 & 0.40 & 10.47 & 8.14 & 3.50 & 5.38 \\
Euclidean & 0.71 & 0.53 & 0.82 & 0.64 & 0.81 & \textbf{0.67} & 10.43 & 8.20 & 3.52 & 5.39 \\
Mahalanobis & 0.36 & 0.33 & \textbf{0.98} & 0.49 & 0.51 & 0.34 & 10.62 & 8.31 & 3.55 & 5.37 \\ \hline

\multicolumn{11}{|c|}{\textbf{B. Pose-specific node removal in GNN (The mentioned Pose = excluded)}} \\ \hline
\rowcolor{gray!20}
None & 0.74 & 0.56 & 0.79 & \textbf{0.66} & \textbf{0.82} & 0.65 & \textbf{6.38} & \textbf{5.32} & \textbf{2.80} & \textbf{2.97} \\
Frontal & 0.71 & 0.57 & 0.77 & 0.36 & 0.44 & 0.55 & 10.74 & 8.29 & 3.55 & 4.02 \\
Lateral & 0.72 & 0.56 & 0.61 & 0.58 & 0.74 & 0.57 & 7.07 & 6.52 & 3.01 & 3.55 \\
Selfie & \textbf{0.76} & \textbf{0.65} & 0.56 & 0.60 & 0.80 & \textbf{0.66} & 6.58 & 5.94 & 3.18 & 5.26 \\
Back & 0.58 & 0.42 & \textbf{0.82} & 0.56 & 0.73 & 0.56 & 6.57 & 6.10 & 3.03 & 4.02 \\ \hline
\end{tabular}
\label{tab:supplementary_ablation}
\end{table*}

\begin{table}[!t]
\centering
\small  % 9pt font allowed for large tables (AAAI guidelines)
\setlength{\tabcolsep}{4pt}  % Adjust column spacing if needed
\setlength{\tabcolsep}{2pt}

\begin{tabular}{lcccccccc}
\toprule
\textbf{Param} & \textbf{Val} & \textbf{F1} & \textbf{AUC} & \textbf{Recall} & \textbf{H} & \textbf{W} & \textbf{MUAC} & \textbf{HC} \\
\midrule
\multicolumn{9}{l}{\textit{Neighbors} ($k$)} \\
$k$ & 3   & 0.6768 & 0.8405 & 0.7667 & 7.24 & 5.32 & 2.44 & 1.94 \\
$k$ & 5   & 0.6700 & 0.8347 & 0.7667 & 7.25 & 5.32 & 2.40 & 1.82 \\
$k$ & 7   & 0.6837 & 0.8363 & 0.7667 & 7.25 & 5.32 & 2.41 & 1.78 \\
$k$ & 10  & 0.6705 & 0.8351 & 0.7667 & 7.25 & 5.32 & 2.40 & 1.77 \\
\addlinespace
\multicolumn{9}{l}{\textit{Class Temperature} ($\tau_{\text{class}}$)} \\
$\tau_{\text{class}}$ & 0.30  & 0.6765 & 0.8327 & 0.7667 & --- & --- & --- & --- \\
$\tau_{\text{class}}$ & 0.50  & 0.6700 & 0.8347 & 0.7667 & --- & --- & --- & --- \\
$\tau_{\text{class}}$ & 0.70  & 0.6800 & 0.8352 & 0.7667 & --- & --- & --- & --- \\
\addlinespace
\multicolumn{9}{l}{\textit{Boost Factor} ($\gamma$)} \\
$\gamma$ & 1.00  & 0.6400 & 0.8342 & 0.7444 & --- & --- & --- & --- \\
$\gamma$ & 1.50  & 0.6400 & 0.8347 & 0.7667 & --- & --- & --- & --- \\
$\gamma$ & 2.00  & 0.6400 & 0.8357 & 0.7667 & --- & --- & --- & --- \\
\addlinespace
\multicolumn{9}{l}{\textit{Regression Temperature} ($\tau_{\text{reg}}$)} \\
$\tau_{\text{reg}}$ & 0.05 & --- & --- & --- & 7.25 & 5.32 & 2.40 & 1.82 \\
$\tau_{\text{reg}}$ & 0.10 & --- & --- & --- & 7.24 & 5.32 & 2.40 & 1.82 \\
$\tau_{\text{reg}}$ & 0.20 & --- & --- & --- & 7.24 & 5.32 & 2.40 & 1.82 \\
$\tau_{\text{reg}}$ & 0.50 & --- & --- & --- & 7.24 & 5.32 & 2.40 & 1.82 \\
\bottomrule
\end{tabular}
\caption{Sensitivity analysis of retrieval hyperparameters on validation set. ``---'' indicates metrics unaffected by that hyperparameter. The metric presented for Regression labels is RMSE}
\label{tab:sensitivity_all}
\end{table}

\subsection{Retrieval Module Sensitivity}

Table~\ref{tab:sensitivity_all} demonstrates that the retrieval-augmented model is highly robust to the choice of all core hyperparameters. Varying the number of neighbors ($k$), class fusion temperature ($\tau_\text{class}$), malnourished boost factor ($\gamma$), and regression temperature ($\tau_\text{reg}$) leads to negligible changes in both classification (F1, AUC, Recall) and regression (RMSE) metrics. All metrics remain effectively unchanged across reasonable ranges, confirming that model performance is not reliant on fine-grained tuning. This robustness supports the reliability and generalizability of our retrieval-augmented fusion design.

\section{Statistical Reliability and Significance}
\subsection{Confidence Interval Analysis}

To quantify statistical reliability across folds, 95 \% confidence intervals (CIs) were computed for all metrics using Student’s $t$ distribution ($df = 3$) across fourfold cross-validation in Table~\ref{tab:supplementary_ci}. 
The mean ± standard deviation values reported in the main paper are reproduced here with corresponding 95 \% CIs. 
These intervals capture per-fold variability and confirm the consistency of NutriScreener (Weighted) relative to the baselines.

\begin{table*}[t]
\centering
\caption{95 \% confidence intervals computed across 4-fold CV using Student’s $t$ distribution ($df=3$). Regression targets have the metric RMSE}
\begin{tabular}{|l|c|c|>{\columncolor{gray!20}}c|}
\hline
\rowcolor{gray!20}
\textbf{Metric} & \textbf{DomainAdapt} & \textbf{CLIP+GNN} & \textbf{NutriScreener (Weighted)} \\ \hline
Accuracy ↑  & 0.68 ± 0.02 → [0.66 – 0.70] & 0.76 ± 0.05 → [0.71 – 0.81] & 0.74 ± 0.05 → [0.69 – 0.79] \\
Precision ↑ & 0.63 ± 0.02 → [0.61 – 0.65] & 0.66 ± 0.06 → [0.60 – 0.72] & 0.56 ± 0.02 → [0.55 – 0.61] \\
Recall ↑    & 0.67 ± 0.05 → [0.62 – 0.72] & 0.54 ± 0.08 → [0.46 – 0.62] & 0.79 ± 0.03 → [0.77 – 0.82] \\
F1 ↑        & 0.64 ± 0.03 → [0.61 - 0.67] & 0.59 ± 0.05 → [0.54 – 0.64] & 0.66 ± 0.02 → [0.65 – 0.68] \\
AUC ↑       & 0.55 ± 0.05 → [0.50 – 0.60] & 0.82 ± 0.05 → [0.77 – 0.87] & 0.82 ± 0.02 → [0.80 – 0.84] \\
mAP ↑       & 0.35 ± 0.03 → [0.32 – 0.38] & 0.66 ± 0.05 → [0.61 – 0.71] & 0.65 ± 0.03 → [0.62 – 0.68] \\
Height ↓    & 22.00 ± 1.62 → [20.38 – 23.62] & 7.35 ± 0.86 → [6.49 – 8.21] & 6.38 ± 0.62 → [5.76 – 7.00] \\
Weight ↓    & 12.40 ± 0.54 → [11.86 – 12.94] & 5.80 ± 0.51 → [5.29 – 6.31] & 5.30 ± 0.45 → [4.85 – 5.75] \\
MUAC ↓      & 3.55 ± 0.41 → [3.14 – 3.96] & 3.80 ± 0.54 → [3.26 – 4.34] & 2.80 ± 0.51 → [2.29 – 3.31] \\
HC ↓        & 5.05 ± 0.33 → [4.72 – 5.38] & 5.23 ± 0.40 → [4.83 – 5.63] & 2.98 ± 0.60 → [2.38 – 3.58] \\ \hline
\end{tabular}
\label{tab:supplementary_ci}
\end{table*}

\subsection{Statistical Significance and Effect Size Analysis}

To assess whether the observed performance differences among models were meaningful, 
we conducted a two–stage analysis: 
(i) a Friedman test to evaluate global inter-model differences across folds, and 
(ii) pairwise comparisons between the proposed NutriScreener (Weighted) model and both 
baselines (CLIP+GNN and DomainAdapt). 
Given the small number of cross-validation folds ($n=4$), formal significance testing 
has limited statistical power; therefore, we additionally report mean fold-wise 
differences ($\Delta$) and Cohen’s $d$, which provide more stable and interpretable 
estimates of effect size under small-sample settings. 
All results are summarized in Table~\ref{tab:effect_size_stats}.

\paragraph{Interpretation.}
The Friedman tests revealed significant overall differences among the three models 
($p<0.05$) for all metrics, indicating that the models produce distinct performance 
profiles. 
Across pairwise comparisons, NutriScreener exhibited the most consistent improvements 
on the core classification metrics, particularly Recall and F1, achieving large to 
extremely large effect sizes (e.g., $d>2$ for Recall and AUC). 
These gains were directionally stable across all folds, suggesting that NutriScreener 
is reliably more sensitive to malnutrition cases while maintaining competitive 
precision and ranking ability.

Compared with CLIP+GNN, NutriScreener showed substantial improvements in Recall and F1, 
and comparable AUC, reflecting a more screening-oriented operating point. 
Compared with DomainAdapt, NutriScreener demonstrated markedly better performance on 
all classification metrics (large effect sizes, $|d|>2$), whereas DomainAdapt remained 
strong on anthropometric regression metrics (Height and Weight), which serve as 
secondary downstream predictors.

Given the limited power inherent to $n=4$ folds, the emphasis is placed on effect sizes 
and directional consistency rather than strict $p$-values. 
Overall, the results support that NutriScreener provides the most balanced and 
clinically meaningful improvements for malnutrition screening, with strong and 
consistent gains on the metrics most relevant to early-risk identification.

\begin{table*}[t]
\centering
\caption{Effect size–based comparison across 4-fold CV. 
Friedman test assesses overall inter-model differences. 
Pairwise comparisons show mean fold-wise differences ($\Delta$) and Cohen's $d$ 
rather than relying only on $p$-values, which have low power at $n=4$. 
Positive $\Delta$ indicates NutriScreener performs better; 
negative indicates baseline better. Regression targets have the metric RMSE}
\begin{tabular}{|l|c|c|c|}
\hline
\rowcolor{gray!20}
\textbf{Metric} 
& \textbf{Friedman ($\chi^2$, $p$)} 
& \textbf{NutriScreener vs CLIP+GNN ($\Delta$, $d$)} 
& \textbf{NutriScreener vs DomainAdapt ($\Delta$, $d$)} \\ 
\hline

Acc ↑ & 7.60, \textbf{0.0224} 
& $-0.04$, $d=-1.41$ 
& $+0.06$, $d=2.78$ \\

Prec ↑ & 6.50, \textbf{0.0388} 
& $-0.10$, $d=-2.97$ 
& $+0.07$, $d=2.10$ \\

Rec ↑ & 8.00, \textbf{0.0183} 
& \textbf{$+0.25$, $d=8.49$} 
& \textbf{$+0.12$, $d=7.83$} \\

F1 ↑ & 7.60, \textbf{0.0224} 
& \textbf{$+0.09$, $d=2.86$} 
& $+0.04$, $d=1.22$ \\

AUC ↑ & 6.00, \textbf{0.0498} 
& $0.00$, $d=0.00$ 
& \textbf{$+0.27$, $d=12.50$} \\

mAP ↑ & 7.43, \textbf{0.0244} 
& $-0.03$, $d=-0.87$ 
& \textbf{$+0.31$, $d=36.74$} \\

Height ↓ & 8.00, \textbf{0.0183} 
& $-1.85$, $d=-6.50$ 
& $-15.63$, $d=-15.66$ \\

Weight ↓ & 7.60, \textbf{0.0224} 
& $-1.23$, $d=-1.28$ 
& $-7.10$, $d=-24.12$ \\

MUAC ↓ & 8.00, \textbf{0.0183} 
& $-1.00$, $d=-12.25$ 
& $-0.75$, $d=-7.50$ \\

HC ↓ & 8.00, \textbf{0.0183} 
& $-2.25$, $d=-17.43$ 
& $-2.08$, $d=-12.15$ \\ \hline

\end{tabular}
\label{tab:effect_size_stats}
\end{table*}

\subsection{Calibration.}
The model’s probabilistic predictions demonstrated strong calibration with an Expected Calibration Error (ECE) of 0.06, Maximum Calibration Error (MCE) of 0.26, and a Brier score of 0.16. 
These low error values indicate that predicted probabilities align well with observed outcomes, confirming that the model is neither over- nor under-confident across thresholds. 
The reliability diagram (Figure~\ref{fig:calibration_decision}) shows that the calibration curve remains close to the ideal diagonal across the full probability range.

\subsection{Decision-Curve Analysis.}
\begin{figure}[t]
\centering
\includegraphics[width=0.47\textwidth]{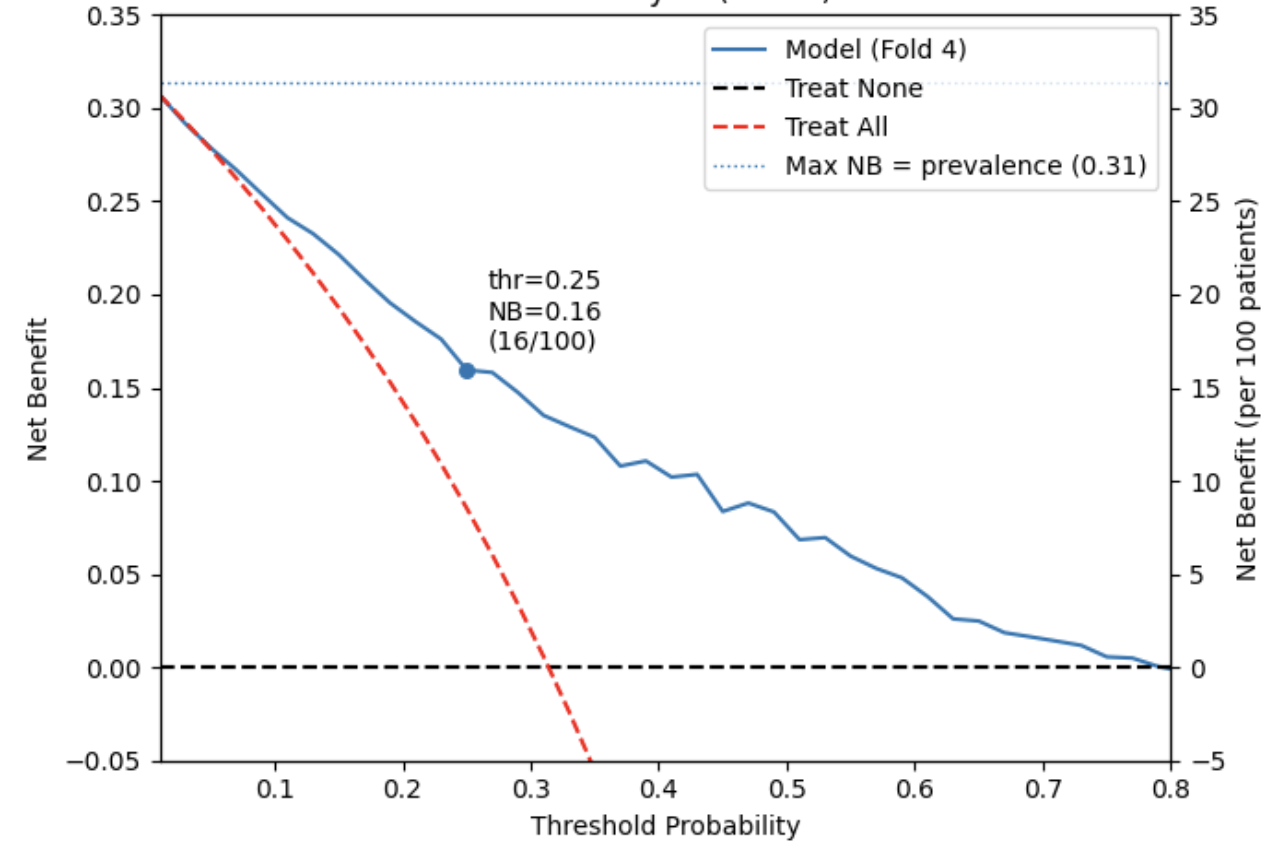}
\caption{
\textbf{Calibration and Decision-Curve Analysis for NutriScreener (Weighted).} 
Calibration plot shows the agreement between predicted probabilities and observed outcomes (ECE = 0.06, MCE = 0.26, Brier = 0.16). 
Decision-curve analysis illustrates net benefit across thresholds: the model achieves maximal NB = 0.16 at threshold $\tau = 0.25$, corresponding to 16 additional correct decisions per 100 patients compared with baseline strategies. 
Dashed red and black lines represent the treat-all and treat-none policies; the dotted blue line marks the theoretical upper bound at the cohort prevalence (0.31).
}
\label{fig:calibration_decision}
\end{figure}

Decision-curve analysis (DCA) quantifies the clinical net benefit (NB) of using the model across decision thresholds ($\tau$) compared with the extremes of treating all or treating none. 
As shown in Figure~\ref{fig:calibration_decision}, the model achieves a maximal net benefit of approximately $+0.15$ at $\tau = 0.3$, corresponding to 15 additional correct decisions per 100 cases relative to the treat-all or treat-none strategies. 
The model’s decision curve remains consistently above both reference lines up to $\tau \approx 0.45$, demonstrating its practical advantage across clinically meaningful probability thresholds. 
The horizontal dotted line indicates the theoretical maximum net benefit corresponding to the cohort prevalence (0.31).

\paragraph{Interpretation.}
Together, the calibration and DCA results confirm that NutriScreener (Weighted) produces well-calibrated probabilities and yields a consistent positive net benefit across realistic decision thresholds. 
This suggests the model is not only statistically superior but also clinically meaningful in guiding population-level screening or triage decisions.

\subsection{Correlation with Knowledge-Base Density}

To understand how the retrieval density influences fusion dynamics, we analyzed the correlation between the learned fusion weight ($\alpha$) and the mean retrieval distance ($\bar{d}$) across all nodes and folds. 
A strong positive correlation ($r = 0.58$, $p < 0.001$) was observed, indicating that denser knowledge-base neighborhoods (i.e., lower $\bar{d}$ values) correspond to higher fusion weights. 
This suggests that the model adaptively emphasizes retrieved representations when the retrieved neighborhood is semantically compact, confirming the intended retrieval-augmented learning behavior.

% \begin{figure}[t]
% \centering
% \includegraphics[width=0.45\textwidth]{CameraReady/LaTeX/images/alpha_vs_density.png}
% \caption{
% \textbf{Correlation between fusion weight ($\alpha$) and knowledge-base density.}
% Each point represents a node’s mean retrieval distance ($\bar{d}$) and corresponding learned fusion weight. 
% A significant positive correlation ($r = 0.58$, $p < 0.001$) indicates that higher KB density (lower $\bar{d}$) is associated with larger $\alpha$, implying greater reliance on retrieved representations in dense semantic neighborhoods.
% }
% \label{fig:alpha_kb_density}
% \end{figure}

% \section{Demo Video}
% A video demonstration of our framework is available for reviewers:\newline
% \url{https://shorturl.at/WaHdZ}\footnote{\url{https://shorturl.at/WaHdZ} Link to anonymized demo video. Last accessed: August 1, 2025.}

\begin{table}[!t]
\centering
\small  % Minimum allowed size for tables
\setlength{\tabcolsep}{5pt}
\begin{tabular}{lccc}
\toprule
\textbf{Metric} & \textbf{Low-Tier} & \textbf{Mid-Tier} & \textbf{High-Tier} \\
\midrule
Processor Model            & i3-12100F  & i5-12400F  & i7-14700K \\
RAM Available              & 4 GB       & 8 GB       & 16 GB \\
\addlinespace
App Launch Time (s)        & 45         & 22         & 8 \\
Image-to-Result Time (s)   & 35         & 12         & 2 \\
Total Screening Time (s)   & 450        & 300        & 240 \\
Peak RAM Usage (MB)        & 822.1      & 354.8      & 203.6 \\
Average CPU Usage (\%)     & 10         & 0.5        & 0.2 \\
\addlinespace
Executable Size (MB)       & \multicolumn{3}{c}{232 MB (shared)} \\
Total Disk Usage (MB)      & \multicolumn{3}{c}{1396 MB (with models and assets)} \\
Offline Capability         & \multicolumn{3}{c}{Fully supported; no internet required} \\
\bottomrule
\end{tabular}
\caption{System performance metrics across CPU-only devices of varying tiers. All results were measured using a compiled standalone executable on CPU-only systems running Windows 10/11.}
\label{tab:sys_perf}
\end{table}

\subsection{System and Deployment Evaluation}

We evaluated the real-world performance of our malnutrition screening application across three CPU-only systems representing low, mid, and high-end deployment scenarios. The goal was to ensure that the system runs efficiently even on low-resource hardware, which is often encountered in field conditions.

The evaluation was conducted on three machines: (1) a low-tier laptop with an Intel i3 processor and 4 GB RAM, (2) a mid-tier laptop with an Intel i5 processor and 8 GB RAM, and (3) a high-end laptop with an Intel i7 processor and 16 GB RAM. All devices were tested without any dedicated GPU acceleration. The app was compiled into a standalone executable (\texttt{.exe}) and run offline without internet access.

Table~\ref{tab:sys_perf} summarizes key performance metrics, including application launch time, average end-to-end inference time per subject (including image loading, CLIP feature extraction, FAISS-based retrieval, and GNN-based classification/regression), peak RAM usage, and average CPU load during execution. Despite the varying hardware capabilities, the system remained responsive and delivered screening results within acceptable latency limits across all tiers.

Notably, even the low-end device completed full subject screening in under \textbf{450} seconds with memory usage under \textbf{822} MB, demonstrating the deployability of our system in resource-constrained settings. The application's offline compatibility further enhances its suitability for remote and underserved regions where connectivity may be limited or unavailable.

\begin{figure}[!t]
    \centering
    \includegraphics[width=\linewidth]{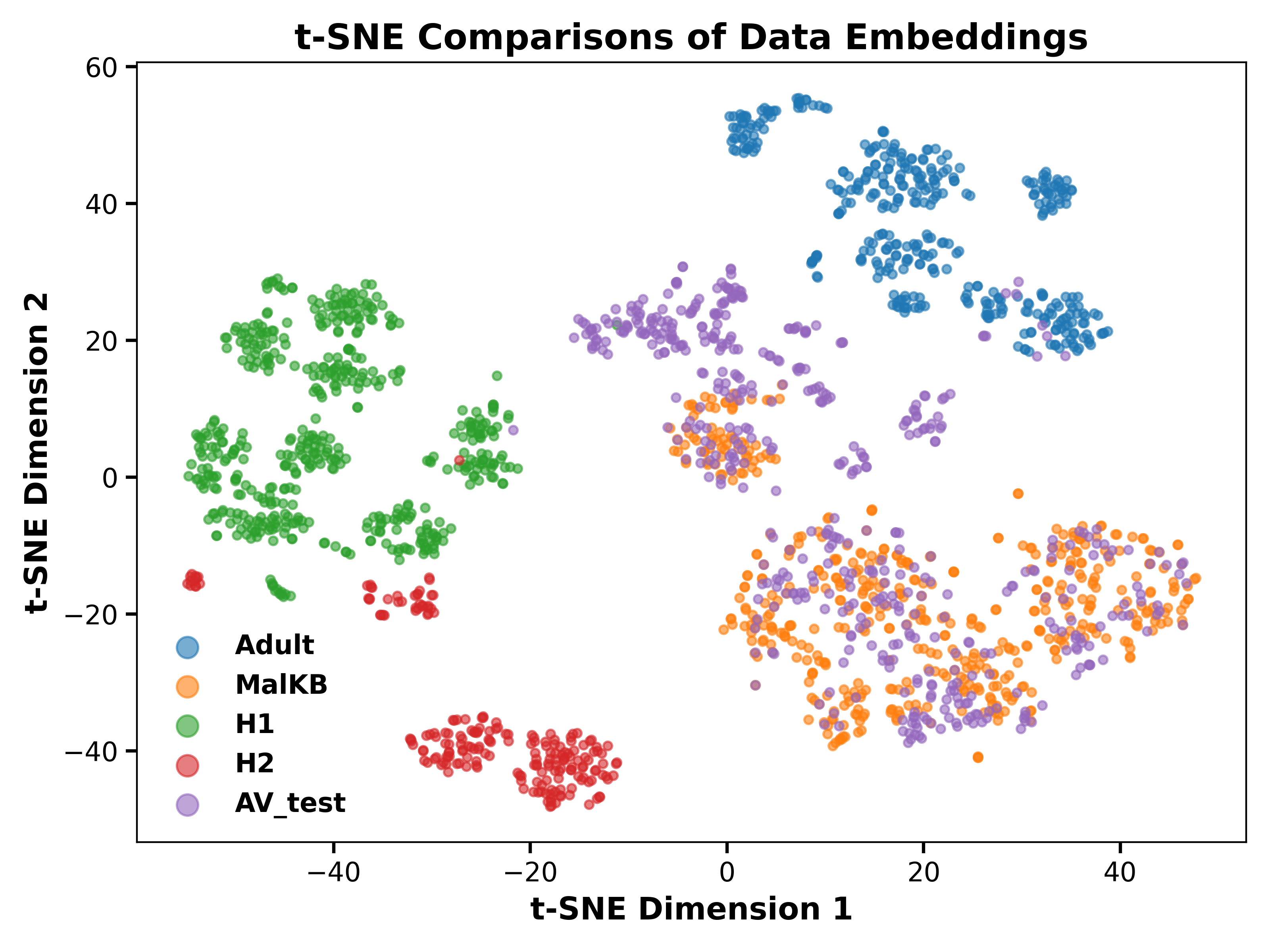}
    \caption{t-SNE of global embeddings showing Adult (blue), MalKB (orange), H1 (green), H2 (red), and AV test (purple). The close overlap of MalKB and AV clusters explains its superior retrieval performance.}
    \label{fig:tsne}
\end{figure}

% \subsection{}
\section{Dataset Demographics and Bias Analysis}

% Table 1: Demographic and Anthropometric Summary
\begin{table*}[t]
\centering
\small  % 9pt font allowed by AAAI for tables
\setlength{\tabcolsep}{5pt}
\begin{tabular}{lcccc}
\toprule
\textbf{Dataset} & \textbf{Height (cm)} & \textbf{Weight (kg)} & \textbf{MUAC (cm)} & \textbf{HC (cm)} \\
\midrule
AnthroVision & $136.83 \pm 19.4$ & $30.76 \pm 12.59$ & $18.72 \pm 4.24$ & $50.87 \pm 3.49$ \\
MalKB        & $134.27 \pm 18.32$ & $29.62 \pm 11.66$ & $17.38 \pm 2.76$ & $50.15 \pm 2.0$ \\
ARAN         & $110.99 \pm 12.2$  & $18.98 \pm 5.18$  & ---              & $50.19 \pm 2.46$ \\
Adult        & $170.21 \pm 14.11$ & $70.73 \pm 19.68$ & $28.56 \pm 8.6$  & $55.98 \pm 3.58$ \\
\bottomrule
\end{tabular}
\caption{Demographic and anthropometric summary of datasets. Values are represented as mean $\pm$ standard deviation.}
\label{tab:anthro_summary}
\end{table*}

% Table 2: Overall Class and Gender Distribution
\begin{table*}[t]
\centering
\small
\setlength{\tabcolsep}{4pt}
\begin{tabular}{lccccccc}
\toprule
\textbf{Dataset} & \textbf{Healthy} & \textbf{Malnourished} & \textbf{\% Mal.} & \textbf{Female} & \textbf{Male} & \textbf{\% Female} & \textbf{Total} \\
\midrule
AnthroVision & 1500 & 641 & 29.94\% & 786 & 1355 & 36.71\% & 2141 \\
MalKB        & 154  & 94  & 37.90\% & 94  & 154  & 37.90\% & 248 \\
ARAN         & ---  & --- & ---     & 258 & 254  & 50.39\% & 512 \\
Adult        & 37   & 40  & 51.95\% & 23  & 54   & 29.87\% & 77 \\
\bottomrule
\end{tabular}
\caption{Overall class and gender distribution across datasets.}
\label{tab:overall_dist}
\end{table*}

% Table 3: Gender-Specific Class Distribution
\begin{table}[t]
\centering
\small
\setlength{\tabcolsep}{5pt}
\begin{tabular}{llcccc}
\toprule
\textbf{Dataset} & \textbf{Gender} & \textbf{Healthy} & \textbf{Maln.} & \textbf{Total} & \textbf{\% Mal.} \\
\midrule
\multirow{2}{*}{AnthroVision} & Female & 536 & 250 & 786 & 31.81\% \\
                              & Male   & 964 & 391 & 1355 & 28.86\% \\
\midrule
\multirow{2}{*}{MalKB} & Female & 60 & 34 & 94 & 36.17\% \\
                       & Male   & 94 & 60 & 154 & 38.96\% \\
\midrule
\multirow{2}{*}{Adult} & Female & 14 & 9 & 23 & 39.13\% \\
                       & Male   & 23 & 31 & 54 & 57.41\% \\
\bottomrule
\end{tabular}
\caption{Gender-specific class distribution, showing percentage of malnourished individuals within each gender group per dataset.}
\label{tab:gender_specific_dist}
\end{table}

To train and evaluate our proposed model, we compiled a comprehensive and diverse dataset from four distinct sources: AnthroVision~\cite{khan2024domainadapt}, MalKB(in-house), ARAN~\cite{MohammedKhan2025ARAN}, and a supplementary Adult(in-house) dataset. This multi-source approach ensures variability in age, ethnicity, health status, and imaging conditions, providing a robust foundation for developing a generalizable model. In total, our collection comprises data from 2,978 unique subjects.

% \subsection{Data Sources and Composition}

% A detailed summary of the anthropometric characteristics for each dataset is presented in Table~\ref{tab:anthro_summary}. The \textbf{AnthroVision} and \textbf{MalKB} datasets feature pediatric subjects with comparable average anthropometry. In contrast, the \textbf{ARAN} dataset contains data from a cohort of younger or shorter children, while the \textbf{Adult} dataset provides an out-of-distribution baseline with adult measurements. This diversity is crucial for testing the model's robustness.

% The overall class (Healthy vs. Malnourished) and gender distributions are detailed in Table~\ref{tab:overall_dist}. The AnthroVision dataset, our largest cohort with 2,141 subjects, has a malnourishment prevalence of 29.94\%. The MalKB dataset exhibits a higher prevalence at 37.90\%. The gender balance varies across sources, with a male majority in the AnthroVision and Adult datasets, while the ARAN dataset is almost perfectly balanced.

% For a more granular analysis, Table~\ref{tab:gender_specific_dist} breaks down the prevalence of malnourishment within each gender. This view reveals interesting trends, such as the higher prevalence of malnourishment or undernutrition in males (57.41\%) compared to females (39.13\%) within the Adult dataset, while the pediatric datasets show a more balanced distribution between genders.

\subsection{Experimental Usage}

In our experiments, the datasets serve distinct roles to ensure a rigorous evaluation protocol. The large-scale \textbf{AnthroVision} dataset serves as the primary source for training, validation, and testing of our main model, conducted via 4-fold cross-validation. The \textbf{MalKB} dataset is leveraged exclusively to construct the external Knowledge Base (KB) that powers our retrieval-augmented algorithm, ensuring a strict separation between the model's learned knowledge and its retrieved knowledge. The \textbf{ARAN} and \textbf{Adult} datasets are reserved for further analysis to evaluate the model's generalization capabilities on related but distinct populations.

\begin{figure*}[!t]
    \centering
    \includegraphics[width=1\linewidth]{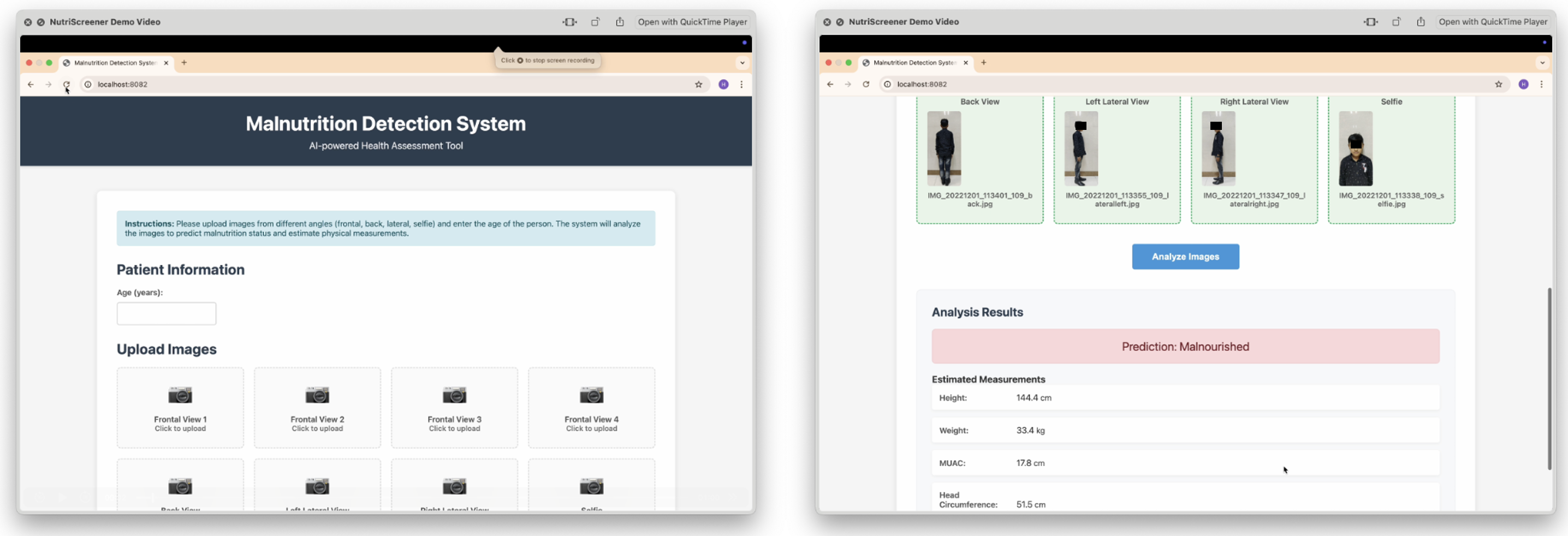}
    \caption{Screenshot of \ourwork{} toolkit}
    \label{fig:demotoolkit}
\end{figure*}

\subsection{Dataset Demographics and Composition}

The study utilizes a composition of four distinct datasets: AnthroVision, MalKB, ARAN, and a smaller Adult dataset for broader context. A comprehensive summary of the anthropometric measurements for each dataset, including mean and standard deviation, is presented in Table~\ref{tab:anthro_summary}. The AnthroVision dataset is the largest, comprising 2,141 subjects, followed by ARAN (512 subjects), MalKB (248 subjects), and the Adult dataset (77 subjects).

The overall class and gender distributions are detailed in Table~\ref{tab:overall_dist}. The AnthroVision and MalKB datasets primarily consist of pediatric subjects with a notable prevalence of malnourishment, at 29.94\% and 37.90\% respectively. The gender distribution varies, with the ARAN dataset being the most balanced at 50.39\% female participants. A more granular, gender-specific breakdown of the class distributions is provided in Table~\ref{tab:gender_specific_dist}, highlighting the varying rates of malnourishment between male and female subjects across different datasets.

\section{User Study: Clinician Feedback}

To assess the real-world applicability and perceived clinical utility of our \ourwork{} toolkit, we conducted a user study with 12 medical professionals, including pediatricians and general practitioners with a range of clinical experience (Mean experience = 9.5 years). The study was designed to evaluate the tool's accuracy, trustworthiness, efficiency, and deployment readiness from an expert's perspective in a live clinical setting. The study began with a brief demonstration video, as shown in Figure~\ref{fig:demotoolkit}, to familiarize participants with the \ourwork{} interface and functionality. Following this, participants were provided with a standalone, executable software version of the \ourwork{} toolkit. It is critical to note that this toolkit contained the trained model and its knowledge base as embedded, non-reversible data structures, ensuring no patient data from the training set was ever revealed or shared. Following this, they were instructed to use the \ourwork{} toolkit on their own patients as part of their regular clinical workflow; on average, all clinicians tested the toolkit on 15 pediatric patients. After applying the tool in this real-world context, participants provided their feedback by completing a questionnaire composed of Likert-scale and open-ended questions. The responses from the Likert-scale questions demonstrate a strong positive reception of the toolkit. The tool's ability to provide classifications consistent with expert clinical judgment received a high mean score of 4.3 out of 5. Participants expressed significant confidence in the system's potential to improve workflow, rating its Efficiency at 4.6 out of 5 (``This tool could help reduce the time and resources required...''). Crucially, trust in the system as a screening and triage aid was rated favorably at 4.4 out of 5. The tool's readiness for deployment with community health workers also scored well, with an average rating of 4.1, suggesting a strong belief in its practical applicability.

Feedback from the open-ended questions provided deeper insights. A recurring theme was the tool's value as an objective ``second opinion,'' particularly in visually ambiguous cases where clinical judgment might be uncertain. One pediatrician noted, ``The tool was particularly impressive in flagging a case of moderate malnourishment that was not immediately obvious from the images alone.'' 

Suggestions for future improvements were consistent among participants, with many requesting the integration of ``uncertainty scores'' to accompany predictions and the ability to ``highlight key visual areas'' that most influenced the model's decision. This feedback validates the tool's core utility and provides a clear roadmap for future development.

Based on the user study, we conclude that: (a) medical professionals perceive \ourwork's predictions as clinically reasonable and consistent with their own expertise; (b) the tool is trusted as a reliable aid that could enhance efficiency in clinical screening workflows; and (c) the system shows significant promise for real-world deployment, especially in empowering community health workers in resource-constrained settings.

\end{document}